\documentclass[11pt]{amsart}

\usepackage{graphicx}
\usepackage{verbatim}
\usepackage{latexsym,amsmath,amsthm,amsfonts,float}
\usepackage[psamsfonts]{amssymb}

\newtheorem{theorem}{Theorem}[section]
\newtheorem{corollary}[theorem]{Corollary}
\newtheorem{lemma}[theorem]{Lemma}

\theoremstyle{definition}

\theoremstyle{remark}

\numberwithin{equation}{section}

\def\GG{\mathbb G}
\def\RR{\mathbb R}

\def\EE{\EXP}
\def\PP{\PROB}

\def\wh{\widehat}

\def\RR{\mathbb R}
\def\EE{\mathbb{E}}
\def\PP{\mathbb{P}}

\def\argmax{\mathop{\rm arg\, max}}
\def\argmin{\mathop{\rm arg\, min}}

\usepackage{graphicx}
\graphicspath{{pics/}}
\DeclareGraphicsExtensions{.eps}

\begin{document}
\title[Dimension reduction  in case control studies]{Dimension reduction and variable selection in case control studies\\ via 
regularized likelihood optimization}
\author[Bunea and  Barbu]
{Florentina Bunea$^{\dag, 1}$}
\address{Florentina Bunea\\ Department of Statistics,   Florida State
University,   Tallahassee,   Florida.} \email{\rm flori@stat.fsu.edu}
\author[]{Adrian Barbu }\address{Adrian Barbu \\ Department of Statistics,   Florida State
University,   Tallahassee,   Florida.} \email{\rm abarbu@stat.fsu.edu}
\thanks{$^\dag$Research partially supported by NSF grant DMS 0406049}
\footnotetext[1] { Corresponding author. }



\date{\today}

\bibliographystyle{amsplain}

\begin{abstract}
Dimension reduction and variable selection are performed routinely in case-control studies, but the literature on the theoretical aspects of the resulting estimates is scarce.  We bring our  contribution to this literature by studying estimators obtained via $\ell_1$ penalized likelihood optimization.  We show that the optimizers of the $\ell_1$ penalized 
retrospective likelihood coincide with the optimizers of  the $\ell_1$  penalized prospective likelihood. This extends the results of Prentice and Pyke (1979), obtained for non-regularized likelihoods.  We establish both the sup-norm consistency of the odds ratio, after model selection, and the consistency of subset selection of our estimators. The novelty of our theoretical results consists in the study of these properties under the case-control sampling scheme. Our results hold for selection performed over a large collection of candidate variables, with cardinality allowed to depend and be greater than the sample size. We complement our theoretical results with  a novel approach of determining  data driven tuning parameters, based on the bisection method.  The resulting procedure  offers significant computational savings when compared with  grid search based methods.  All our numerical experiments support strongly our theoretical findings.

\end{abstract}
\maketitle

\section{Introduction}

Case-control studies investigate the relationship between a random outcome $Y$, typically the disease status, and a number of candidate variables $X_j$, $ 1 \leq j \leq M$,  that are potentially associated with $Y$.  An important instance is provided by 
cancer studies, where the $X_j$'s may quantify exposures to certain substances, or may be a collection of genes or genetic markers. One of the  problems of interest in such studies is the  identification, on the basis of the observed data,  of a smaller subset of  the set of candidate variables, that can reliably suffice for subsequent analyses. This problem becomes more challenging when  the collection of potential disease factors $M$ is very large.  Our solution to this problem is variable selection via penalized likelihood optimization.   We propose below a computationally efficient selection method and  we  investigate the theoretical properties of our estimates under the case-control data generating mechanism.  We begin by giving the formal framework used throughout the paper and by making our objectives precise.

We consider   binary  outcomes $Y\in \{0, 1\}$, where $Y = 0$ labels the controls (non-disease), and  $Y = 1$ labels the cases (disease). Let $f_0(x), f_1(x)$ be, respectively,  the conditional distributions of $X = (X_1, \ldots, X_M)$ given $Y = 0$ and $Y = 1$. 
Under the case-control or retrospective sampling scheme we observe two independent samples  from each of  these conditional distributions. Formally, we observe
\begin{eqnarray}\label{retros} X_1^0\ldots, X_{n_{0}}^{0} \  \mbox{i.i.d. with density} \ f_0(x), \\
 X_1^1\ldots, X_{n_{1}}^{1} \ \mbox{i.i.d. with density} \ f_1(x), \nonumber \end{eqnarray}
\noindent where the superscripts 0 and 1 of $X$ are mnemonic of the fact that the samples correspond to $Y = 0$ and $Y = 1$, respectively. For simplicity of notation we assume that $n_0 = n_1 = n$.  We assume that the outcome $Y$ is connected to $X$ via  the logistic link function 
\begin{equation}\label{pi}
 P(Y = 1|X=x)  = \frac{\exp{( \delta_0 + \beta^{\prime}x)}}{ 1 + \exp (\delta_0 +  \beta^{\prime}x)},\end{equation}
\noindent where $\beta \in \RR^M$, $\delta_0 \in \RR$.  \\

\noindent  In this article we will further assume that $\beta = \beta^*$, where $\beta^* \in \RR^M$   has non-zero components corresponding to an index set $I^*\subseteq \{1, \ldots, M\}$ with cardinality $|I^*| = k^*$,  possibly much smaller than  $M$.  The variable selection  problem can be therefore rephrased in this context as the problem of estimating the unknown set $I^*$ from data generated as in (\ref{retros}).  We note that this problem is not equivalent
with the problem of estimating $I^*$ from i.i.d. pairs $(X_i, Y_i)$, with $Y_i$ generated from (\ref{pi}), as no random samples  from the distribution of  $Y$ are available under the sampling scheme  (\ref{retros}). However,  the likelihoods corresponding to the two sampling schemes are intimately related, with results detailing their connections dating back to the 1970s. This link is essential for our procedure and we illustrate it below. Following Prentice and Pyke (1979), Section 3,  an application of the  Bayes' theorem combined with rearranging  terms gives the following re-parametrization of $f_0$ and $f_1$, respectively:
\begin{eqnarray} \label{crux} f_0(x) &= & 2\times \frac{ 1}{ 1 + \exp(\delta + \beta^{\prime}x)}\times q(x)=: 2p_0(x)q(x), \\
f_1(x) &=& 2\times \frac{\exp(\delta+ \beta^{\prime}x)}{ 1  + \exp(\delta + \beta^{\prime}x)}\times q(x)=: 2p_1(x)q(x),\nonumber \end{eqnarray}

\noindent where  $ \delta$ is a  new intercept parameter, different than $\delta_0$,  $\beta$ is given by (\ref{pi}), and $q(x)$ is a positive function that integrates to one.  The parameters $\delta$, $ \beta$ and $q$ are constrained by the requirement that $f_0$ and $f_1$ are probability densities, that is \begin{equation}\label{constraint}\int p_j(x)q(x)dx = \frac{1}{2}; \ \ j \in \{0, 1\}. \end{equation} 

Therefore, the  likelihood function corresponding to  data generated as in (\ref{retros}), to which we will refer in the sequel  as to the  {\it retrospective likelihood}, is:  

\begin{eqnarray}\label{defl}
 L_{retro}(\delta, \beta, q) &=:& \Pi_{i=1}^{n}f_0(x_i^0)f_1(x_i^1) \\
 &=& \left \{ \Pi_{i=1}^{n}p_0(x_i^0)p_1(x_i^1)  \right \} \times \left \{ 4\Pi_{i=1}^{n}q(x_i^0)q(x_i^1)\right \} \nonumber \\
&=: & L_{pros}(\delta, \beta) \times L(q),\nonumber \end{eqnarray}
\noindent with parameters $\delta, \beta$ and $q$ related via the constraint (\ref{constraint}). Notice that  $ L_{pros}$ is,  up to the intercept, exactly the standard logistic regression likelihood,  had we observed $2n$ i.i.d observations  $(X_i, Y_i)$, with equal number of 0 and 1 responses $Y_i$ 
generated according to  (\ref{pi}); this quantity and the corresponding sampling scheme are typically referred to in this context as  the {\it prospective likelihood}, and {\it prospective sampling scheme}, respectively. We will also use this terminology below.   \\

The earlier results on the estimation of $\beta$ via (\ref{defl}) did not address the model selection problem and were mostly concerned with the asymptotic properties of the estimates of  $\beta$.  Anderson (1972),  Prentice and Pyke (1979), Farewell (1979)  are among the pioneers of  this  work and showed that: \\

\noindent (1) The vector $\tilde{\beta} \in \RR^M$ that maximizes the retrospective likelihood 
$L_{retro}(\delta, \beta) $  under the constraint (\ref{constraint}) coincides with  $\bar{\beta} \in \RR^M$ that maximizes the prospective likelihood  $L_{pros}(\delta, \beta)$; \\

\noindent  (2) The asymptotic distribution of  $\tilde{\beta}$, derived under the sampling scheme (\ref{retros}) coincides with the asymptotic distribution of $\bar{\beta}$ derived under the 
prospective sampling scheme. \\

A number of important works continued this program, and 
provided in depth analyses of various other aspects of the estimators of $\beta$ in retrospective studies. We refer to  Gill, Vardi and Wellner (1988) and Carroll, Wang and Wang (1995), for more general sampling schemes, to Qin and Zhang (1997) for goodness of fit tests, to Breslow, Robins and Wellner (2000) for a study of the efficiency of the estimators,  to Murphy and van der Vaart (2001), for partially observed covariates,  to Osius (2009), for general semiparametric association models and to Chen, Chatterjee and Carroll (2009),  for shrinkage methods tailored to inference in haplotype-based case-control studies and  the asymptotic distribution of the resulting estimators. The variable selection problem was not considered in any of these works.  Although model selection techniques are routinely used in case-control studies, and are typically based on testing via the asymptotic distribution of $\widehat{\beta}$,  we are unaware of 
theoretical analyses of the performance of the resulting estimators under this sampling scheme. \\  

Our contribution to this literature is to provide answers to the model selection analogues of (1), and to formulate goals that replace (2) above by goals targeted to the dimension reduction and selection aspects. Specifically, we propose a model selection method based on a penalized likelihood approach, with a sparsity inducing penalty. In this article we will focus on the $\ell_1$ penalized likelihood with tuning parameter $\lambda$.   We will  show the following: \\

\noindent (I) For {\it any}  penalty function $ pen(\beta) $ that is independent of $\delta$,  the maximizer of 
$L_{retro}(\delta, \beta) + pen(\beta)$ under the constraint (\ref{constraint}) coincides with the  maximizer of  the prospective likelihood  $L_{pros}(\delta, \beta) + pen (\beta)$.  \\

\noindent (II) For $pen(\beta) = \lambda\sum_{j = 1}^{M}|\beta_j|$,  we obtain estimators $\hat{I}$ of $I^*$ and dimension reduced estimators $\widehat{\beta}$ of $\beta^*$ by optimizing  $L_{pros}(\delta, \beta) + pen (\beta)$. Then:\\

(a) The behavior of $\PP(\widehat{I} = I^*)$, analyzed under the sampling scheme (\ref{retros}) is essentially the same as  the behavior of  $\PP(\widehat{I} = I^*)$, evaluated under the prospective sampling schemes. \\

(b) The  estimator $\widehat{\beta}$, analyzed under the sampling scheme (\ref{retros}),  adapts to the unknown sparsity of $\beta^*$, which  parallels the same property that can be established for $\widehat{\beta}$, under the prospective sampling schemes. \\

\noindent The result announced in (I) is an immediate extension of result (1) above
established in the 1970s for the unpenalized likelihood. We present it  in Section 2 below. The results in (II) necessitate an analysis that is  completely different  than the one needed for (2), and we discuss (a) in Section 4 and (b) in Section 3.  The immediate implication of (b)  that is relevant to case-control  studies is the fact that  the estimator $\widehat{\beta}$, which is supported on a space of potentially much lower dimension  than the original $\RR^M$, yields sup-norm consistent estimates of the odds ratio, where the odds ratio is defined as follows.  The odds of having $Y=1$ for an individual with characteristics $X = x$ are $ O(x) = P(Y = 1|X=x)/P(Y=0|X=x)$, and the odds ratio is defined as $O(x)/O(x_0)$, for some reference characteristic  $X = x_0$.  Under model (\ref{pi}), the odds ratio becomes
\begin{equation}\label{eq:oddsratio} R(x) =: \exp(\beta^{'}(x - x_0)).\end{equation}
We establish the after model selection consistency  of estimators of this ratio  in Section 3 below. \\

In this article we will concentrate on the analysis of estimators obtained by optimizing $\ell_1$ penalized criteria. The literature on the theoretical aspects of such estimates  has seen an explosion in the past few years, together with the development of  efficient algorithms for computing them.  The results pertaining to generalized linear models are most closely connected to our work,  and  all of them have  been established for what we termed above the prospective sampling scheme, that is for  data consisting of $(X_i, Y_i)$ i.i.d. pairs. For analyses conducted under this framework, we refer to van der Geer (2008) and Bunea (2008),  for  sparsity oracle inequalities for the Lasso,  and to Bunea (2008), for correct subset selection;  we refer to  Ravikumar,  Wainwright and  Lafferty (2008) for related results on graphical models for binary variables.   Motivated by the increasing usage of the Lasso type estimators for the analysis of data arising from case-control studies, see e.g.,  Shi, Lee and Wahba (2007),  Wu et al (2009) and the  references therein,  we complement the existing literature on this type of estimators by providing their theoretical analysis under the case-control data  generating mechanism (\ref{retros}).  To the best of our knowledge, this is the first such analysis 
proposed in the literature. \\

The rest of the paper is organized as follow.  Section 5  complements the theoretical results of Sections 2 - 4, by  providing  a fast algorithm for finding a data driven tuning parameter  for the $\ell_1$ penalized optimization problem.  	Our procedure does not use a grid search. Instead, we use a generalization of the bisection method to compute 
tuning parameters that yield, respectively, estimators with exactly $ k = 0, 1, \dots, M$ non-zero components. We then use  a 10-fold cross-validated logistic loss function to which we add a BIC-type penalty to select one of these estimators. The corresponding procedure is easy to implement and offers important computational savings over a grid search based procedure, which can be 50 times  slower,  for the same degree of accuracy.  Section 6 contains 
a detailed analysis of the proposed estimators, via simulations. It supports very strongly all our theoretical findings. Section 7 is a conclusion section, summarizing our findings. All the proofs are collected in the Appendix.

\section{Penalized logistic regression  for case-control data}

In this section we investigate the type of penalty functions for which estimation of $\beta$ via   penalized {\it retrospective} log-likelihood optimization reduces to the estimation of $\beta$ via  penalized {\it prospective}  likelihood optimization. Recall that we mentioned  in (\ref{defl}) above that the  retrospective likelihood can be written as the product of the prospective likelihood and a term depending on $q$:
\begin{eqnarray} L_{retro}(\delta, \beta, q) &=: & L_{pros}(\delta, \beta) \times L(q), \nonumber \end{eqnarray}
\noindent where the parameters $\delta$, $ \beta$ and $q$ are constrained by (\ref{constraint}). Let $pen(\beta)$ be a function that depends on $\beta$ alone, and is independent of $\delta$ and $q$.  Define the unconstrained maxima
\begin{equation}\label{unc} (\widehat{\delta}, \widehat{\beta}) = \argmax_{\delta, \beta}\left \{\log L_{pros}(\delta, \beta)+ pen(\beta)\right \};  \ \ \ \widehat{q} = \argmax_{q} \log L(q),
\end{equation}
\noindent where the second maximum is taken over all density functions $q$. The following result, proved in the Appendix, is an immediate extension of the result derived by  Prentice and Pike (1979) for the unpenalized likelihood.  
\begin{lemma}\label{samelik}
Let $pen(\beta)$ be any function independent of $\delta$ and $q$. Let $\widehat{\delta}, \widehat{\beta}$ and $\widehat{q}$ be given by (\ref{unc}). Then  \[((\widehat{\delta}, \widehat{\beta}), \widehat{q}) = \argmax_{\delta, \beta, q} \{\log L_{retro}(\delta, \beta, q) + pen (\beta)\},\] 
where the maximizer is computed over all $\delta, \beta, q$ satisfying the constraint (\ref{constraint}).
\end{lemma}

\noindent Lemma \ref{samelik} shows that the penalized log-likelihood estimates of  $(\delta, \beta$), for a likelihood corresponding to data generated as in (\ref{retros}) coincide with the penalized prospective log-likelihood estimates, which we rescale by $2n$: 
\begin{eqnarray}\label{log}  
(\widehat{\delta}, \widehat{\beta}) &=&  \argmin_{\delta, \beta} \{ - \frac{1}{2n}\log L_{pros}(\delta, \beta) + pen(\beta) \}  \\
&=:&  \argmin_{\delta, \beta} ( \frac{1}{2n}\sum_{i=1}^{n}\log( 1 + e^{\delta + \beta^{\prime}{X}_i^1} ) - \frac{1}{2n}\sum_{i=1}^{n}(\delta +  \beta^{\prime}{ X}_i^1) \nonumber \\
&& + \frac{1}{2n}\sum_{i=1}^{n}\log ( 1 + e^{\delta + \beta^{\prime}{ X}_i^0}) + pen(\beta)).
      \nonumber \end{eqnarray}

  Lemma \ref{samelik} holds for any function $pen(\beta)$, as long as it is independent of $\delta$ and $q$. Since we are interested in dimension reduction, we will consider a sparsity inducing penalty. Throughout this paper our estimates will be obtained via (\ref{log})  with the  $\ell_1$ penalty given below 
\begin{equation}\label{pen}
pen(\beta) = \lambda\sum_{j=1}^{M}|\beta_j|,
\end{equation}
\noindent for a tuning parameter $\lambda$ that will be made precise in the following  section.

\section{Consistent estimation of the odds ratio \\ after variable  selection }

\noindent 

 Recall that the true odds ratio (\ref{eq:oddsratio}) is given 
in terms of $\beta^*$, which  is supported on a space of dimension $k^*$,  possibly much smaller than $M$.  We show that the estimated odds ratio based on the selected variables corresponding to the non-zero elements of $\widehat{\beta}$ given by (\ref{log}) above, for penalty (\ref{pen}),  provides a consistent estimate, in the supremum norm, of the odds ratio: 
\begin{equation}\label{want} \sup_x |\exp\widehat{\beta}^{\prime}(x - x_0)  - \exp{\beta^*}^{\prime}(x - x_0)| \longrightarrow 0,\end{equation}
with probability converging to one. For simplicity of notation, we assume in what follows that $x_0 = 0$. For uniformity of notation, we also denote the intercept parameter given in   (\ref{crux})  by $\delta^*$.

 Our arguments are
based on the following central fact, that may be of independent interest.   Define the functions 
 \begin{eqnarray} \label{ls}  \ell_0(\theta) &= :&   \ell_0(\theta; x) \  = \ \log ( 1 + e^{\theta^{\prime}x}), \nonumber \\
\ell_1(\theta) &=: & \ell_1( \theta; x) \ = \ \log ( 1 + e^{\theta^{\prime}x}) - \theta^{\prime}x,\nonumber \end{eqnarray} 
\noindent and recall the notation $\theta = (\delta, \beta)$.  In order to write what follows in a  compact way we further denote by  $\PP^0$ and $\PP^1$ the probability measures corresponding to the densities $f_0$ and $f_1$ defined in (\ref{crux}), respectively.  For a generic function $g$  we write  $ \PP^0g$ and $\PP^1g$ for integration with respect to  $\PP^0$ and $\PP^1$,  respectively. 
With this notation we define the quantity 
 \begin{eqnarray}\label{deltadif}\Delta(\widehat{\theta}, \theta^*) = \frac{1}{2} \PP^0 \left( \ell_0 (\widehat{\theta})  - \ell_0(\theta^*) \right)    +   
\frac{1}{2}\PP^1 \left(    \ell_1  (\widehat{\theta}) -   \ell_1(\theta^*)  \right).
\end{eqnarray}

\noindent Theorem \ref{delta}  below, which is central to our paper, establishes that the difference $\Delta$ is small. The proof, given in the Appendix,  uses the control of appropriately scaled empirical processes corresponding to the two samples. If  $\Delta\left((\widehat{\delta}, \widehat{\beta}), (\delta^*, \beta^*)\right)$ is small with high probability, relatively standard arguments can be used to show that \[ \sup_{x}|\widehat{\beta}^{\prime}x -  \beta^{*\prime}x  + \widehat{\delta} - \delta^*|\] is also small, with high probability, and we establish this in Corollary \ref{firstsup} below. 

To arrive at our desired result (\ref{want}),  we need to complement Corollary \ref{firstsup} with  a study of the difference $|\widehat{\delta} - \delta^*|$. This is done in Theorem \ref{rates}, which also contains a stronger result: it shows that $\widehat{\beta}$ adapts to the unknown sparsity of $\beta^*$, in that it is a consistent estimator of $\widehat{\beta}$, with the rate of convergence of an estimate based only on $I^*$ variables.

 The combination of Theorem \ref{delta},  Corollary \ref{firstsup} and Theorem \ref{rates} 
yields  the desired  sup-norm consistency of the odds ratio stated in  Theorem \ref{thm:oddsratio}.  All  proofs are collected in the Appendix. All our theoretical results will be proved under the assumption that the design variables and the true parameter components are bounded. We formalize this in Assumption 1 below. \\

\noindent {\it Assumption 1.} \\

\noindent  (i) There exits a constant $L > 0$, independent of $M$ and $n$, such that $|X_i^j| \leq L$, for all $i$ and $j$, with probability 1. \\

\noindent (ii)  There exits a constant $B >0$, independent of $M$ and $n$, such that $\max_j|\beta_j^*| \leq B$; $|\delta^*| \leq B.$ \\

 \noindent Let $\delta_n$ be any sequence converging to zero with $n$. Define the tuning sequence 
\begin{equation}\label{er} r =\log n  \left(6L \sqrt{\frac{2\log 2 (M \vee n)}{n}} + \frac{1}{  4(M \vee  n)}  +  4L\sqrt{\frac{2\log\frac{1}{\delta_n}}{n}}  \right). \end{equation}

\noindent If $\delta_n = 1/n$ and $M$ is polynomial in $n$, the tuning sequence $r$ is 
of the order $\frac{\log n \sqrt{\log n}}{\sqrt{n}}$.  The results of this section will be relative to estimators obtained  via the penalty (\ref{pen}),  with tuning parameter $\lambda = 2r$. For compactness of notation, we let $\widehat{\theta} =: (\widehat{\delta}, \widehat{\beta}) \in \RR^{M +1}$, \ $\theta^* =: (\delta^*, \beta^*) \in \RR^{M + 1}$. Whenever we use this compact notation, we will also use the notation $ \widehat{\theta}^{\prime}u$ or 
${\theta^{*}}^{\prime}u$, for some vector $u \in \RR^{M + 1}$ that will be  implicitly assumed to have the first component equal to 1.

\begin{theorem} \label{delta} Under Assumption 1,  if $k^*r \rightarrow 0$, then  for any $\alpha > 0$ 
\[ \PP (\Delta(\widehat{\theta}, \theta^*) > \alpha ) \longrightarrow 0,\]
as $n \rightarrow \infty$.
\end{theorem}

\noindent We give  an immediate corollary of this theorem which establishes the sup-norm consistency of $\widehat{\theta}^{\prime}x$. It is interesting to note that both Theorem \ref{delta} above and the corollary below hold under no assumptions on the dependence structure of the design variables. 

\begin{corollary}\label{firstsup} Under {\it Assumption 1},  if $k^*r \rightarrow 0$, then for any $\gamma > 0$
\[ \PP\left(\sup_{x}|(\widehat{\beta}^{\prime}{x} - {\beta}^{*\prime}{x}) + ( \widehat{\delta} - \delta^*)| \geq \gamma\right) \longrightarrow 0,\]
as $n \rightarrow \infty$.

\end{corollary}

\noindent The following theorem establishes rates of convergence for the estimates of $\delta^*$ and $\beta^*$.  It requires minimal conditions on the design variables. We formalize them below.  Let $V$ be a $(M + 1) \times (M + 1)$ matrix containing a $(k^* + 1) \times (k^* + 1)$ identity matrix, corresponding to $I^*$ and the intercept,  and with zero elements otherwise.  Let $X_1, \ldots, X_n$ denote $X_1^0, \ldots, X_n^0$, and let  $X_{n+1}, \ldots, X_{2n}$ denote $X_1^1, \ldots, X_n^1$. By convention, each $X_i$ is viewed as a vector in $\RR^{M + 1}$ with the first component equal to 1, i.e. $X_{0i}  = 1$, for all $i$. We used this compact notation to avoid unnecessary superscripts. Let $\Sigma$ be the $(M + 1) \times (M + 1)$  sample Hessian matrix  with entries \[ \frac{1}{2n}\sum_{i=1}^{2n}p_0(X_i)p_1(X_i)X_{ki}X_{ji}, \  0 \leq j, k \leq M. \] 

\medskip 

\noindent {\it Condition  H}. There exists  $ 0 < b \leq 1$ such that $\PP(\Sigma - bV \geq 0) = 1$. \\

\noindent {\it Remark.}  {\it Condition H}  is a mild condition on $\Sigma$, as it requires that  this matrix remain semi-positive definite after a slight modification of some of its diagonal elements, those corresponding to $I^*$. This relaxes the conditions needed
for the  consistency of the estimators based on the non-regularized log-likelihood,  when one requires that the Hessian matrix be positive definite, see e.g. Prentice and Pyke (1979).\\

Let $c = 6/bw$, for $b$ given by   {\it Condition H} above and for some positive number $w$ that is arbitrarily close to 1.
\begin{theorem}\label{rates} Under {\it Assumption 1} and {\it Condition H},  if $k^*r \rightarrow 0$, then 
\begin{eqnarray}
&& (i) \ \ \PP\biggl(|\widehat{\delta} - \delta^*| \leq   crk^*\biggr) \longrightarrow 1,
\nonumber \\ 
&& (ii) \ \ \PP\biggl (\sum_{j=1}^{M}|\widehat{\beta}_j - \beta_j^*| \leq  crk^*\biggr ) \longrightarrow 1, \nonumber 
\end{eqnarray}
\noindent as $n \rightarrow \infty$.
\end{theorem}

\noindent

The theorem shows that the rate of convergence of $\widehat{\beta}$ adapts to the unknown sparsity of $\beta^*$: $\sum_{j=1}^{M}|\widehat{\beta}_j - \beta_j^*|$ has $M$ terms,  so we expect its size to be equal to the optimal rate $1/\sqrt{n}$ of each term multiplied by $M$. Theorem \ref{rates} shows that in fact $\widehat{\beta}$ behaves like an estimator obtained in dimension $k^*$, had this dimension been known, as its rate of convergence in the $\ell_1$ norm is, up to constants and logarithmic terms, $1/\sqrt{n}$ multiplied by $k^*$, for our choice of $r$.  \\

\noindent Theorem \ref{rates}  is the result announced in II(b) of the introduction: the estimators of $\beta^*$ analyzed under the retrospective sampling scheme exhibit the same adaptation to sparsity as those analyzed under the prospective sampling scheme.  For a full analysis of the $\ell_1$ penalized logistic regression estimates based on i.i.d. $(X_i, Y_i)$ pairs, with $Y_i$ generated as in (\ref{pi}) we refer to Bunea (2008), for results obtained under  conditions similar to {\it Condition H} on the Hessian matrix,  and  to van de Geer (2008), for results on generalized linear models obtained under conditions on the covariance matrix of the covariates. The difference between the results obtained under the two sampling schemes is essentially minor, and  consists in a slight difference in the size of the tuning parameter $r$, which needs to be larger, by a $\log n$ factor, for the case control studies. This  is a very small price to pay for not having full information on the joint distribution of $X$ and $Y$. \\

 \noindent  Corollary \ref{firstsup} and  Theorem \ref{rates}  {\em (i)} immediately imply, via a first order Taylor expansion,  the desired result of this section. We summarize  it in  Theorem \ref{thm:oddsratio} below which shows that  for appropriate choices of the tuning parameters $r$, and if the size of the true model does not grow very fast relative to $1/r$ then, under minimal conditions on the covariates,  we can estimate the odds ratio consistently. 
 \begin{theorem}\label{thm:oddsratio} Under {\it Assumption 1} and {\it Condition H},  if $k^*r \rightarrow 0$, then 

 \[ \PP(\sup_{x}|\hat{R}(x) - R(x)| \leq \alpha) \longrightarrow 1,\]
 for any $\alpha > 0$, as $n \rightarrow \infty$. 
 \end{theorem}

 \section{Consistent variable selection}

In this section we investigate the consistency of  the index set $\widehat{I}$ corresponding to the non-zero components of  the estimator $\widehat{\beta}$ discussed above. We show that  $ \PP( I^* = \hat{I}) \longrightarrow 1$  holds for  the retrospective scheme, under conditions similar to those needed in  the prospective sampling scheme. We state them below. \\

\noindent {\it Condition 1}. \ There exists $d >0$ such that 
\[  \PP\Bigg( \max_{ j \in I^*, k \neq j }\Bigg| \frac{1}{2n}\sum_{i=1}^{2n} X_{ij} X_{ik}\Bigg|
\leq \frac{d}{k^*}\Bigg) = 1.\]

\noindent For $p_0$ and $p_1$ defined in (\ref{crux}) above, we assume that the following also holds. \\

\noindent {\it Condition 2}. \ There exits $d >0 $ such that 
\[ \PP\Bigg( \max_{ j \in I^*, k \neq j }\Bigg| \frac{1}{2n}\sum_{i=1}^{2n} p_0(X_i)p_1(X_i)X_{ij} X_{ik}\Bigg|
\leq \frac{d}{k^*}\Bigg) = 1.\]

\noindent {\it Remark 1.} The constants $d$ in the two conditions above need not be the same;  we used the same letter for clarity of notation. 
\noindent {\it Remark 2.} The two conditions above can be regarded as conditions guaranteeing the identifiability of the set of true variables $I^*$. {\it Condition 1}  requires that there exists some degree of separation between the variables in $I^*$ and the rest, in that the correlations between variables in these respective  sets are  bounded, up to constants, by $1/k^*$. If $k^*$ is small to moderate, the restriction is  mild.  {\it Condition 2} reinforces {\it Condition 1}, by requiring that these variables remain separated even when separation is measured by the entries in the Hessian matrix, which can be regarded as weighted  correlations.  \\

\noindent Let $\delta_n$ be any sequence converging to zero with $n$. Define the tuning sequence 
\begin{equation}\label{newr}  r =\log n  \left(6L \sqrt{\frac{2\log 2 (M \vee n)}{n}} + \frac{1}{  4(M \vee  n)}  +  4L\sqrt{\frac{2\log\frac{M}{\delta_n}}{n}} \right), \end{equation}

\noindent and notice that the last term  in this definition of $r$ differs  by a factor of $\sqrt{\log M}$ from the last term  in  $r$ given in (\ref{er}) of  the previous section.  If $\delta_n = 1/n$ and $M$ is polynomial in $n$, the tuning sequence is 
again of the order $\frac{\log n \sqrt{\log n}}{\sqrt{n}}$. The following assumption reflects the fact that we can only detect coefficients above the noise level, as quantified by $r$. \\

\noindent {\it Assumption 2.}  $\min_{j \in I^*} |\beta_j^*| > 4r$.

\begin{theorem}\label{sel} Under Assumptions 1 and 2,  if $k^*r \rightarrow 0$ and {\it Conditions 1} and {2} are met, then 
$\PP( I^* = \hat{I})  \longrightarrow 1,$ as $n \rightarrow \infty$.
\end{theorem}

\noindent Theorem \ref{sel} shows that $\widehat{I}$, analyzed under the retrospective sampling scheme, is a consistent estimator of $I^*$, under essentially the same conditions needed for its analysis under the prospective sampling scheme, see Bunea (2008) and also  Ravikumar,  Wainwright and  Lafferty (2008)  for related results.

\section{Data based tuning sequences and the bisection method} 

In Sections 3 and 4  above we showed  that  if the cardinality of the true model is not larger than $\sqrt{n}$,  up to constants,  the proposed method yields consistent estimation of the odds ratio and of the support of $\beta^*$, for tuning sequences $r$ given in (\ref{er}) and (\ref{newr}), respectively.  Since the constants involved in these expressions may be conservative, we complement our theoretical  results  by offering in this section a fully data driven construction  of the tuning parameters. Typical methods involve two main steps.  In the first step one computes the regularization path  of the solution to (\ref{log}), as the tuning parameter $r$ varies. Then, in a second step, one  selects the appropriate value of $r$ from a fine grid of 
values,  by cross-validating the log-likelihood. \\

 We present here a method that follows in spirit this idea,  but with two main modifications: we do not need to compute the regularization path, but rather a sketch of it and we choose $r$   via a combined cross-validation/BIC -type procedure. We describe the resulting technique in what follows. \\
  
 We begin by noting that, unlike $\ell_1$ penalized least squares, the regularization path for the  $\ell_1$ penalized logistic likelihood is not piecewise linear and it cannot be computed analytically, see e.g.,  Rosset and Zhu (2007), Koh et al.  (2007).  In this case, approximate regularization paths can be obtained from path following algorithms, see, e.g.,  Hastie et al. (2004),  Park and Hastie (2006, 2007),  Rosset (2005). These constructions rely on the following observation:  the values of the tuning parameter $r$ define a partition of the positive axis, such that for all $r$ belonging to  an interval of this partition we obtain the same sparsity pattern.  Thus, the full information on the sparsity pattern can be recovered from having a representative inside each such interval.

\begin{figure}[ht]
\centering
\includegraphics[width=5.5cm]{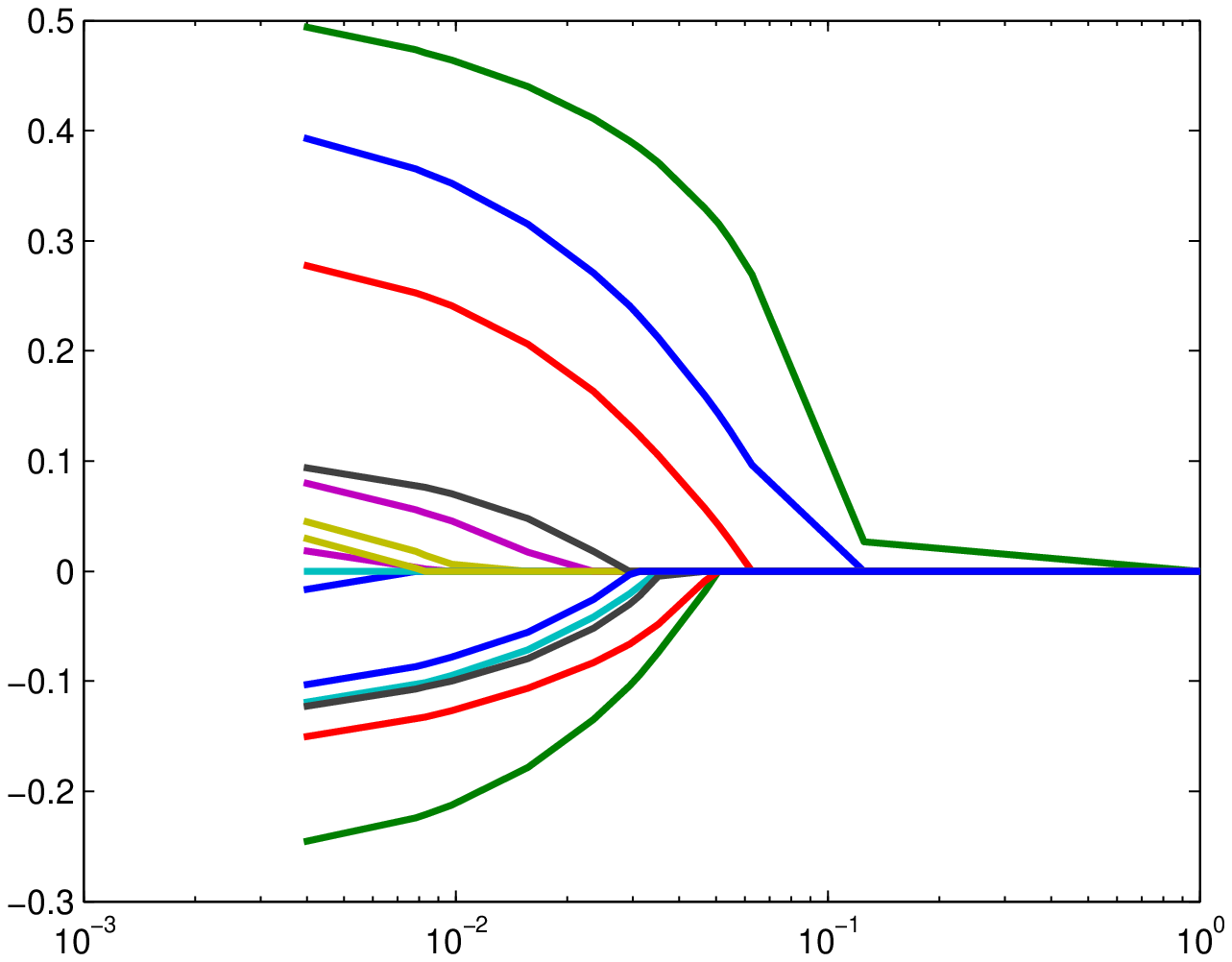}
\includegraphics[width=5.5cm]{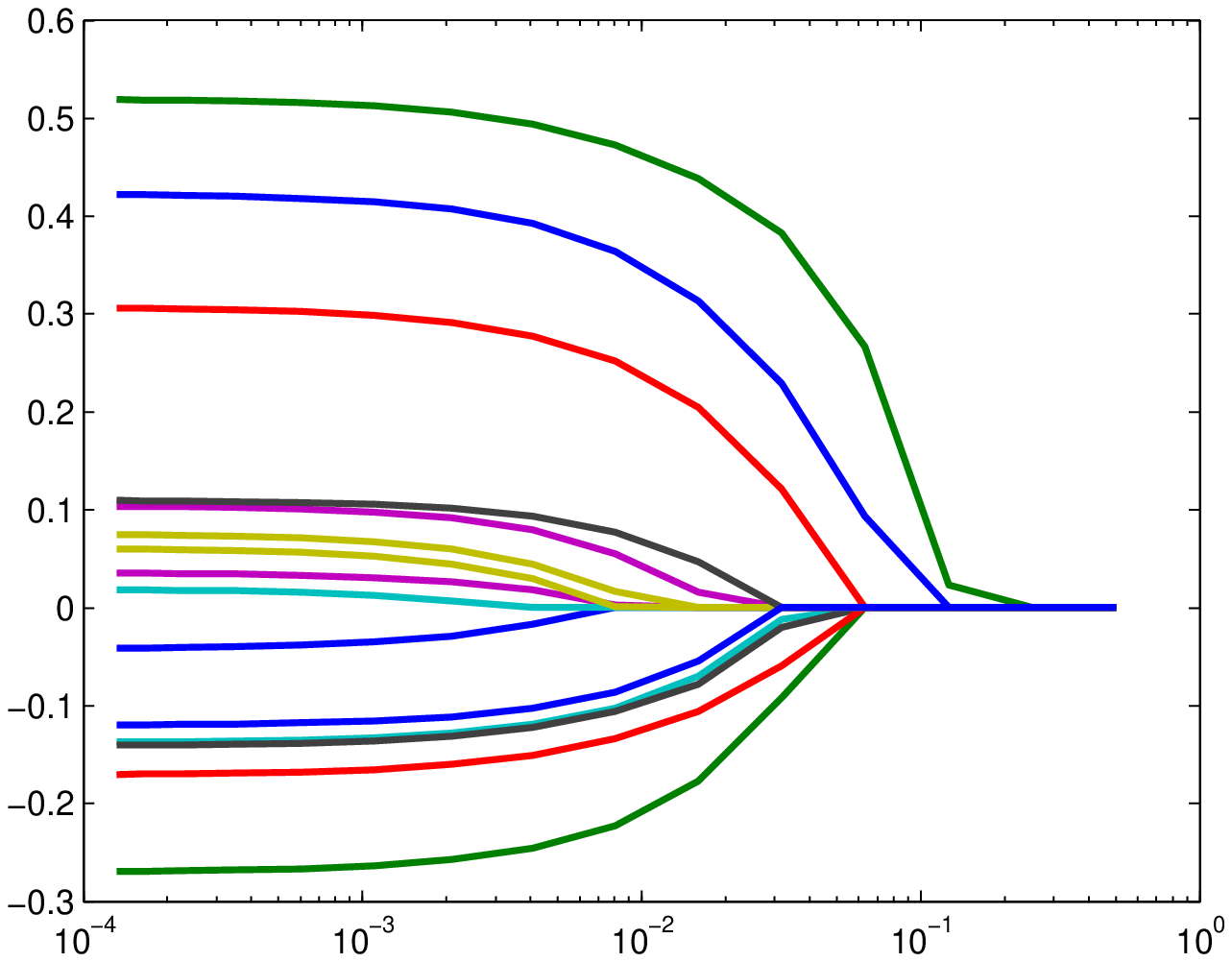}

\caption{Approximate regularization path for normal iid data, $2n=300$, $M=15$ and $k^*=3$. Left: the regularization path sketch of $\widehat{\beta}$, obtained using {\bf GBM}. Right: regularization path of $\widehat{\beta}$ using a grid search with the same computational complexity as the {\bf GBM}.}
\label{fig:sketch}
\end{figure}

  We call  the path corresponding to these representative values  a "sketch" of the regularization path. The problem therefore reduces to finding the set of representatives. For this, one could perform a grid search and find  the intervals  where the sparsity pattern change. However, a grid search method may not include some of these intervals, as they can have arbitrary length. Thus, a grid search  may skip some of the sparsity patterns that are in fact present in the regularization path.  Figure \ref{fig:sketch} below offers an instance of this fact. 

\vskip -3mm

\noindent It shows that a  grid search with the same computational complexity as  the generalized bisection method {\bf GBM}, described in detail  below, can fail to contain  the true index set $I^*$ in the corresponding regularization path. In this example $I^*$ has three elements, which can be recovered in the left panel, and not in the right panel, as  there is no value of $r$ in the grid we considered for which we can have exactly three non-zero components.  \\

Our procedure  replaces the grid search with a different method, that allows us to obtain, for each dimension $1 \leq k \leq M$,  a value of $r$ for which the solution given by \eqref{log} has exactly $k$ non-zero components.  For each tuning parameter $r$,  let the number of nonzero entries in $\hat \beta_r$ be denoted by $\hat n(r)$. For each dimension
$1 \leq k \leq M$ we let  $h_k(r) = \hat n(r) - k$. Our method consists in finding $r$ such that $h_k(r) = 0$. To solve this equation we use  the Bisection Method, e.g. Burden and Faires (2001), which is a well established computationally efficient method  for finding a root $z \in \RR$ of a  generic function $h(z)$. The bisection method can be summarized as follows. Let $\alpha$ be the desired degree of accuracy of the algorithm. \\

\noindent {\bf The Basic Bisection Method (BBM).} \\
Given $\alpha>0$, do:
\begin{enumerate}
	\item[1.] Choose $z_0, z_1$ such that $h(z_0)h(z_1)\leq 0$ (i.e. $h(z_0), h(z_1)$ have different signs).
	\item[2.] If $h(z_0)=0$ or $h(z_1)=0$ stop.
	\item[3.] Take $z = (z_0 + z_1)/2$. If $|z_1- z_0| < \alpha$ then stop and return $z$.
	\item[4.] If $h(z)h(z_0) < 0$, make $z_1 = z$. 
	\item[5.] Else $h(z)h(z_1) < 0$ and make $z_0 = z$.
	\item[6.] Return to step 3.
\end{enumerate}

\noindent We can apply the basic bisection method to solve  $h_k(r) = 0$, for each $k$, and  obtain the desired values of the tuning sequence $r_1, \ldots, r_M$. However, performing {\bf BBM} for each dimension $k$ separately is not computationally efficient, as  there is a large amount of overlapped computation. In what follows we propose an extension of the {\bf BBM} that finds a sequence $r_0,...,r_M$ such that $\hat n(r_k)=k$, for all  $0 \leq k\leq M$. The extension uses a queue consisting of pairs $(r_i, r_j)$ such that $\hat n(r_i)<\hat n(r_j) - 1$. \\

\noindent {\bf The General Bisection Method for all $k$ (GBM).}\\
 Initialize all $r_i$ with $-1$.
\begin{enumerate}
	\item[1.] Choose $r_0$ very large, such that $\hat n(r_0) = 0$. Choose $r_n = 0$, hence $\hat n(r_n) = n$.
	\item[2.] Initialize a queue $q$ with the pair $(r_0, r_n)$.
	\item[3.] Pop the first pair $(a, b)$ from the queue.
	\item[4.] Take $r = (a+b)/2$. Compute $k = \hat n(r)$. 
	\item[5.] If $r_k = -1$ make $r_k = r$. 
	\item[6.] If $|\hat n(a) -k|>1$ and $|a-r|>\alpha$, add $(a, r)$ to the back of the queue. 
	\item[7.] If $|\hat n(b )-k|>1$ and $|b-r|>\alpha$, add $(r, b)$ to the back of the queue.
	\item[8.] If the queue is not empty, go to 3.
\end{enumerate}

\medskip 

The {\bf GBM} described above offers a way of  obtaining  candidate values $r_0, \ldots, r_M$ for the tuning parameter $r$. Observe that this method does not use anything specific to the $\ell_1$-regularized logistic loss function.  The {\bf GBM} can be used in connection with any penalized loss function, for a  sparsity inducing penalty. This makes the {\bf GBM} method more general than the one proposed in  Park and Hastie (2007), which is restricted to $\ell_1$- regularized generalized linear models. \\

 We performed a quantitative evaluation of the {\bf GBM} on four problems of different difficulty levels. The data were simulated from a Normal distribution, described in Section 6 below as NOR\_IID.  In Table \ref{tab:grid} below we compared 
 the  {\bf GBM} and a  grid search  in terms of their capability of constructing  approximate regularization paths containing the true $I^*$.  The percentages reported in the table are percentages of time $I^*$ was in the path,  over 250 simulations. Table \ref{tab:grid}  indicates that  if the set $I^*$ is in the regularization path, it will also be in the {\bf GBM} path sketch, obtained at a low computational expense. 

\begin{table}[ht] 
\small
\begin{center}
\begin{tabular}{c c c c c c c c}
Experiment &Size $2n$ &$ k^*=|I^*|$ &GBM &Grid &Grid$\times$10 &Grid$\times$50\\
   \hline
  \hline
1 &100 &3 &70.0 &28.8 &64.8 &70.0\\
2 &200 &3 &99.2 &79.2.7 &99.2 &99.2\\
3 &300 &10 &88.4    &76.8    &87.2    &87.6\\
4 &400 &10 &98.8    &93.2    &98.8    &98.8\\
\hline
\end{tabular}
\vskip 1mm
\caption{Percentage of times the $I^*$ was present in different approximate regularization paths for the Normal dataset NOR\_IID, $M=250$. Grid has the same computational complexity as the {\bf GBM}, while Grid$\times$10 and Grid$\times$50 are finer grids that are $10$ respectively $50$ times more expensive than the {\bf GBM}.\label{tab:grid}
}
\end{center}
\end{table}

 To complete the selection procedure,  we use the  dimension stabilized $p$-fold cross-validation procedure  summarized below. Let $D$ denote the whole data set, and let $D=D_1\cup \ldots \cup D_p$ be a partition of $D$ in $p$ disjoint subsets, each subset containing the same percentage of cases and controls. In all the experiments presented in the next section we took $p = 10$. 
 Let $D_{-j} = D \setminus D_j$.  We will denote by $r_k^j$ a candidate tuning parameter determined using the {\bf GBM} on $D_{-j}$.  We denote by $I_k^j$ the set of indices corresponding to the non-zero coefficients of the estimator of $\beta$ given by (\ref{log}), for tuning parameter $r_k^j$ on $D_{-j}$.  We denote the unpenalized maximum likelihood estimators 
corresponding only to the variables with indices in  $I_k^j$ and computed on $D_{-j}$
by $(\hat{\delta}^j, \hat{\beta}_k^j)$.  With $\log L_{pros}(\delta, \beta)$ defined in  
(\ref{log}) above, let $L_k^j =: \log L_{pros}(\hat{\delta}^j, \hat{\beta}_k^j)$, computed on $D_j$.  With this notation, the procedure becomes: 
\vskip 2mm
\noindent {\bf Variable Selection Procedure.}\\
Given: a dataset $D$ partitioned into $p$ disjoint subsets: $D=D_1\cup \ldots \cup D_p$, each subset containing the same percentage of cases and controls. Let $D_{-j} = D \setminus D_j$ for all $j$.
\begin{enumerate}
\item [1.] For each $ 1 \leq k \leq M$ and each  fold $j$ of the partition, $ 1 \leq j \leq p$:\\
 \indent Use the {\bf GBM} to find $r_k^j$ and $I_k^j$ such  $\hat n(r_k^j) =|I_k^j|=k$ on $D_{-j}$.\\
 \indent Compute $L_k^j=: \log L_{pros}(\hat{\delta}^j, \hat{\beta}_k^j)$, as defined above. 
\item [2.] For each $1 \leq k \leq M$: \\
\indent Compute $L_k =: \frac{1}{p}\sum_{j = 1}^{p}L_k^j$. 
\item [3.] Obtain 
\vspace{-1mm}
\[\hat{k} = \argmin_{k} (L_k + 0.5k \frac{\log 2n}{2n}).\vspace{-1mm}\] 
\item [4.] With $\hat{k}$ from Step 3, use the {\bf BBM} on the whole data set $D$ to find  the tuning sequence $r_{\hat{k}}$ and then compute the final estimators using (\ref{log}) and this tuning sequence. 
\end{enumerate}

\noindent  {\it Remark.} In Section 4 above we showed that the tuning sequence required for correct variable selection needs to be slightly larger than the one needed for accurate estimation of the odds ratio. This suggests  that cross-validation, which is routinely used for the latter purpose, is not the appropriate method for constructing a data driven tuning sequence for accurate variable identification. This fact is becoming well recognized in practice and was also pointed out in Leng, C., Lin, Y., and Wahba, G. (2006), for linear models. This motivates the modification we used in Step 3 of our procedure described above, where we added a BIC-type penalty to the cross-validated loss function. BIC-type methods  have been used successfully for  correct variable identification in a  large variety of contexts, and we refer  to Bunea and McKeague (2005) and the references therein for related results in prospective models. The numerical experiments presented  in Section 6 below indicate very strongly that the same is true for our variable selection procedure, applied to the logistic likelihood and case-control type data.  The theoretical analysis of this procedure is beyond the scope of this paper and will be addressed in future work.

\section{Numerical Experiments}
\subsection{Simulation design}
In this section we illustrate our proposed methodology via simulations, for the  three data generating mechanisms described below. We generate $X = (X_1, \ldots, X_M) \in \RR^M$ according to one of the following three distributions:
\begin{enumerate}
	\item  {\em SNP:} $X_1, \ldots, X_M$ are i.i.d. as $U$, where $U$ is the discrete random variable having zero mean and variance $1$:
\[
U\sim \begin{pmatrix} -\sqrt{2} & 0  & \sqrt{2}\\ 1/4 &1/2 &1/4\end{pmatrix}.\]
	\item {\em NOR\_IID:} $X \sim N(0,\sigma^2I_M)$.
	\item {\em NOR\_CORR:} $ X \sim N(0,\Sigma)$.
\end{enumerate}

\noindent We used the  label {\em SNP} for the first type of $X$ we considered, as this type of distribution is  encountered in the analysis of Single Nucleotide Polymorphism type data;  the i.i.d. assumption is not typically met  for the basic SNP data sets, but is  instead met for what is called tagging SNPs, which are collections of SNP representatives picked at large enough intervals from one another to mimic independence; we used the label SNP here for brevity.

The parameters of the other two distributions are chosen as : $ \sigma^2 = 1$ and $\Sigma = (\rho^{|i-j|})_{i,j}$ with $\rho=0.5$. \\

\noindent For each of the three distributions of $X$ given above we generated two samples, of size $n$ each, from $f_0(x)$ and $f_1(x)$ described in (\ref{retros}), respectively, for given $\PP(Y = 1) = \pi$ and using the logistic link (\ref{pi}). 

The non-zero entries of $\beta$ were set to  $1$, that is  $\beta_j=1, j\in I^*$. In our experiments, we took $\pi=P(Y=1)=0.01$, i.e. we assumed a rare incidence of the disease. The value of  $\delta_0$ given by  (\ref{pi}) was found numerically in order to obtain the desired incidence $\pi$. 

\subsection{The behavior of $\wh \beta'x$}

The quality of the estimators of the odds ratio $R(x)$, for a given baseline $x_0$, is dictated by the quality of  $\wh \beta^{\prime}x$ as an estimator of $\beta^{*\prime}x$. Notice that the sup-norm consistency of  $\wh \beta^{\prime}x$ that follows from 
  Corollary \ref{firstsup} and  Theorem \ref{rates}  {\em (i)} above immediately implies 
  its consistency in  empirical norm or mean squared error $MSE = \frac{1}{2n}\sum_{i=1}^{2n}(\wh \beta^{\prime}X_i  - \beta^{*\prime}X_i)^2$. We present below a numerical study of the  mean squared error {\em MSE} for $k^* = 3$ and $k^* = 10$, for different values of $n$ and $M$, with $M$ possibly larger than $n$. In all the results of this section we report the median {\em MSE}s over 200 simulations. 
\begin{figure}[ht]
\centering
\includegraphics[width=5.5cm]{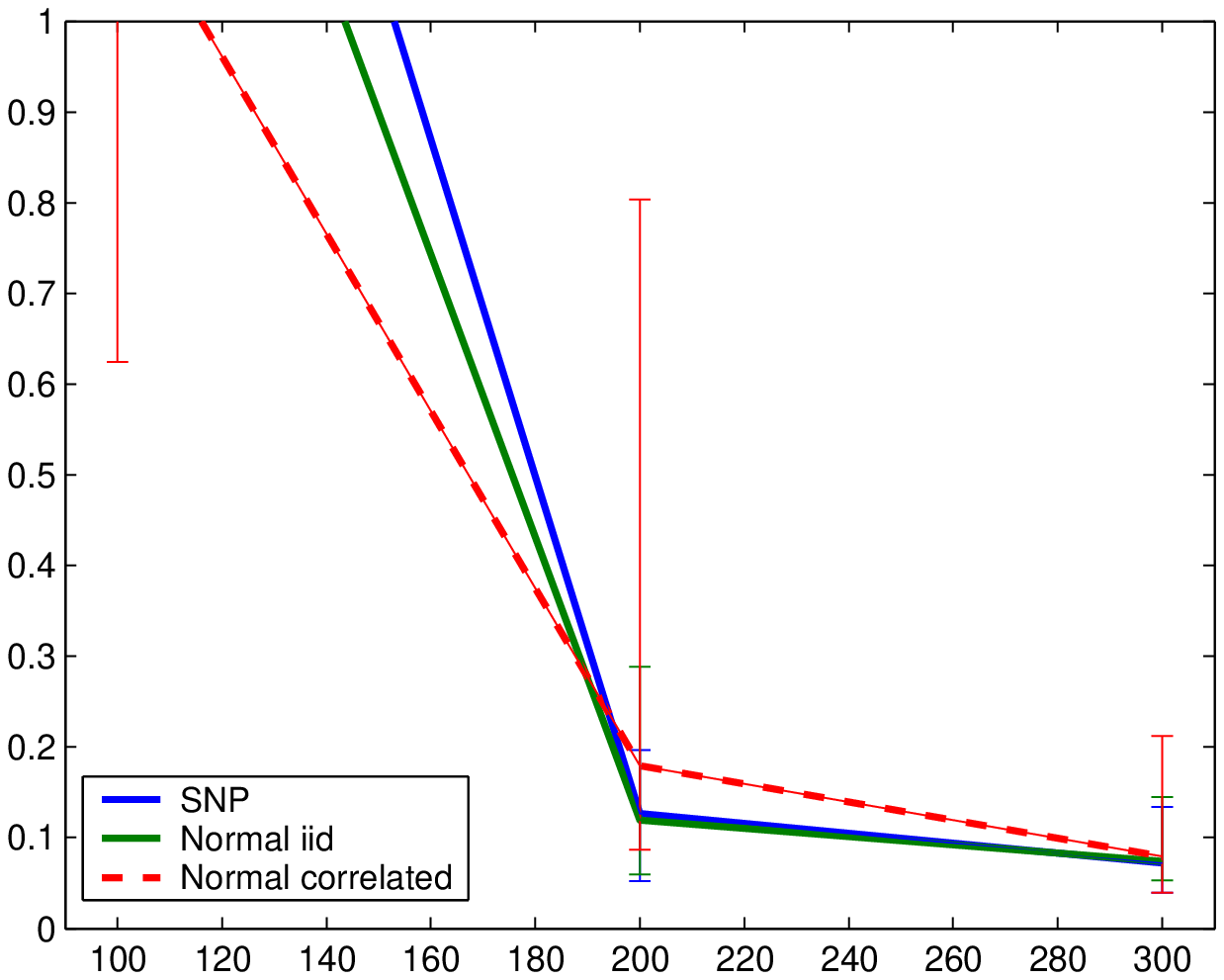}
\includegraphics[width=5.5cm]{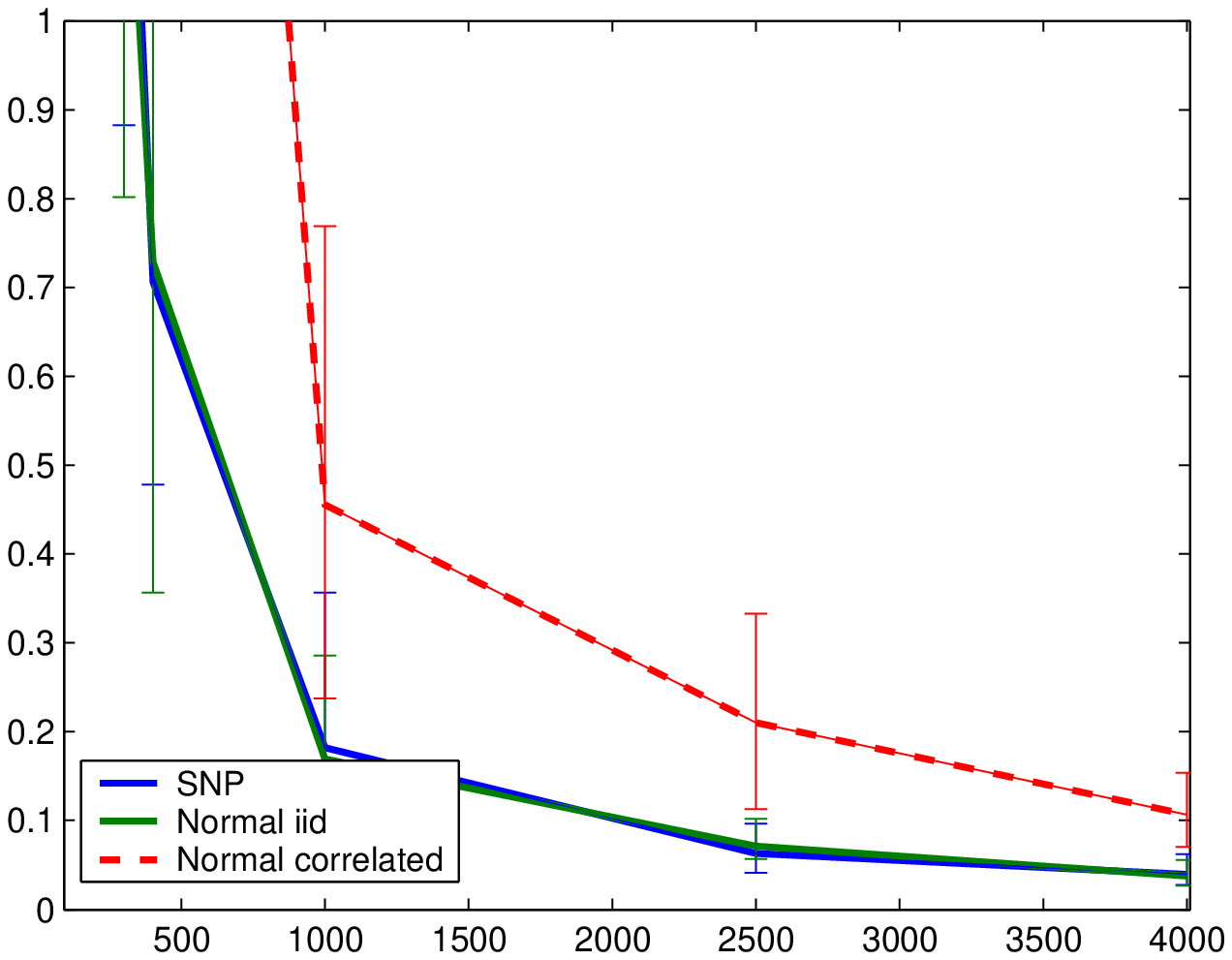}

\caption{The median error vs. number $2n$ of observations, $M = 2000$.
Left:  $k^* = 3$, right: $k^* = 10$. }
\label{fig:R_n}
\end{figure}

\begin{figure}[ht]
\centering
\includegraphics[width=5.5cm]{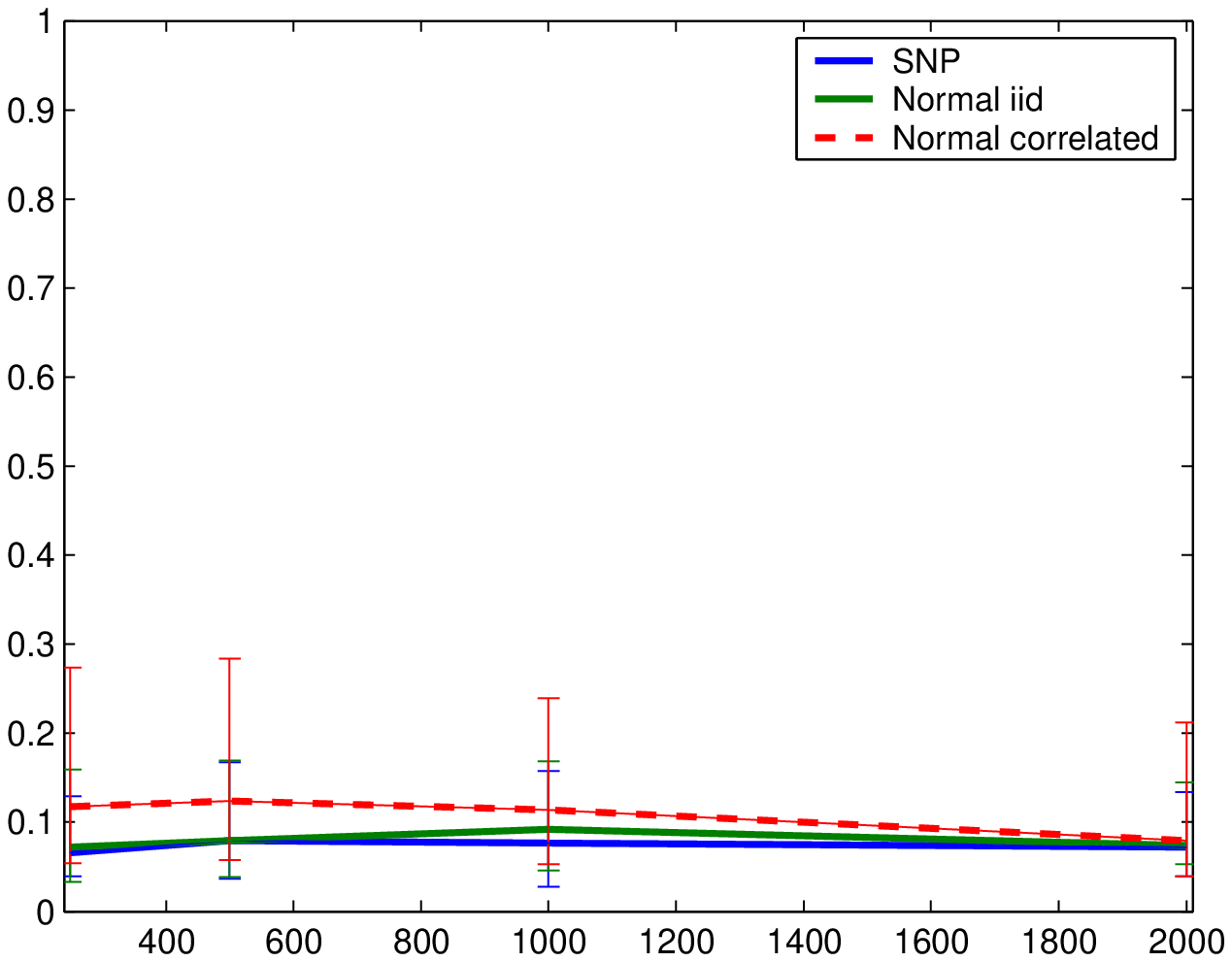}
\includegraphics[width=5.5cm]{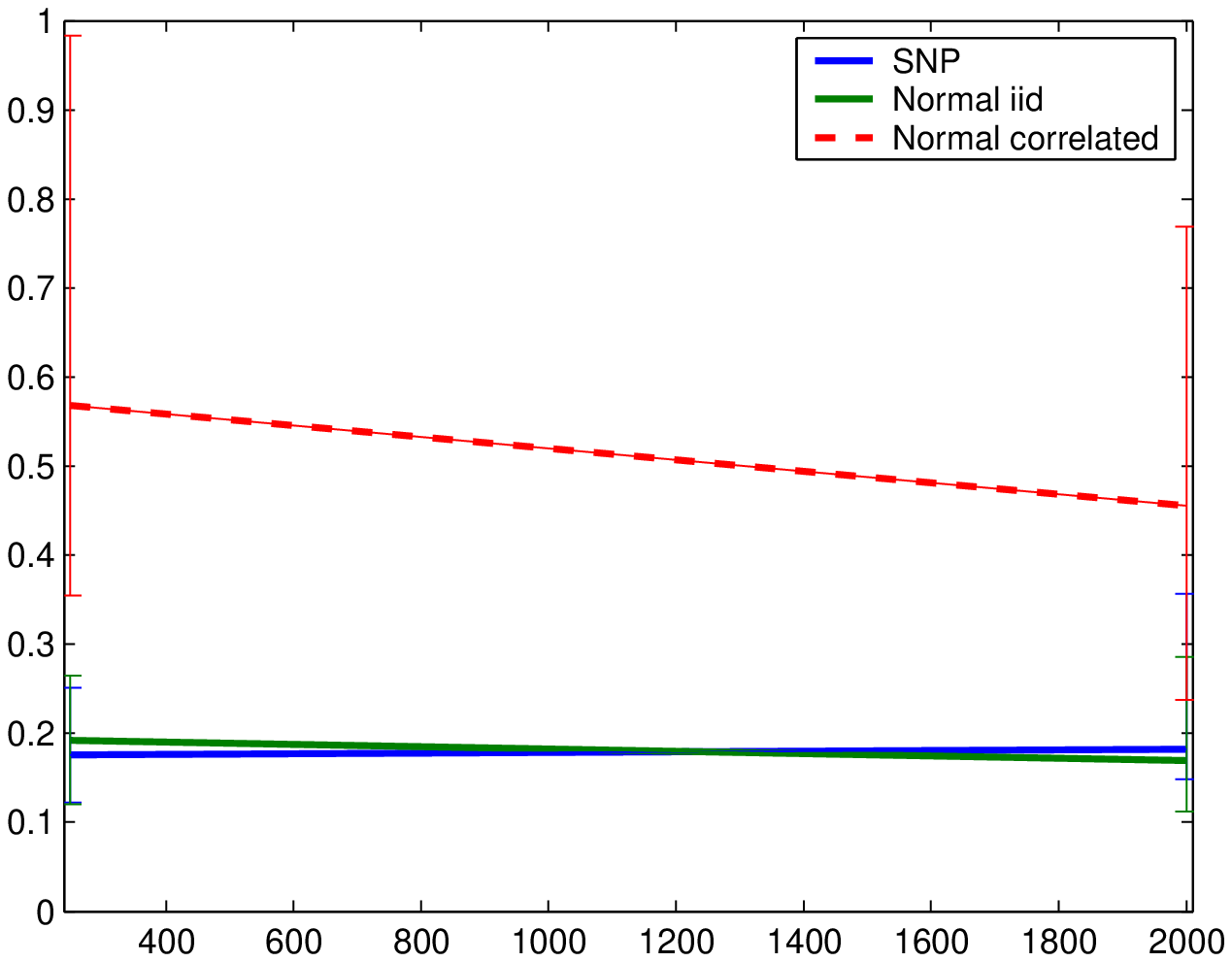}
\vskip -3mm
\caption{The median error vs number $M$ of predictors.  Left: $2n=300, k^*=3$, right: $2n=1000, k^*=10$.
 }
\label{fig:err_m}
\end{figure}

\noindent Figure \ref{fig:R_n} shows the decrease of the {\em MSE} as the sample size increases, when $M$ is kept fixed and set to $M = 2000$ in this example. 
Figure \ref{fig:err_m} shows that for  given configurations $(k^*, 2n)$ the size of the {\em MSE}  is essentially unaffected by an increase in the number of predictors $M$ from 250 to 2000.  

Our findings are consistent with the behavior of the Lasso estimates in other generalized linear models and with our  Theorem \ref{thm:oddsratio} of Section 3.  The overall conclusion is that the size of the {\em MSE}  is influenced by the model size $k^*$ and not by $M$, the total number of variables in the model. When the variables $X_j$ are correlated, the {\em MSE} values are higher for the larger value of $k^* = 10$ and very similar to the {\em MSE} values obtained for  the uncorrelated variables when $k^* = 3$.  This supports the fact that the dependency structure of  the variables $X_j$'s  per se does not have a dramatic  impact on the {\em MSE}, as suggested by our theoretical results of Section 3, which hold under the very mild {\it Condition H} on the design variables. \\

\subsection{Variable selection accuracy} \label{sec:varsel}

\begin{figure}[ht]
\centering
\includegraphics[width=5.5cm]{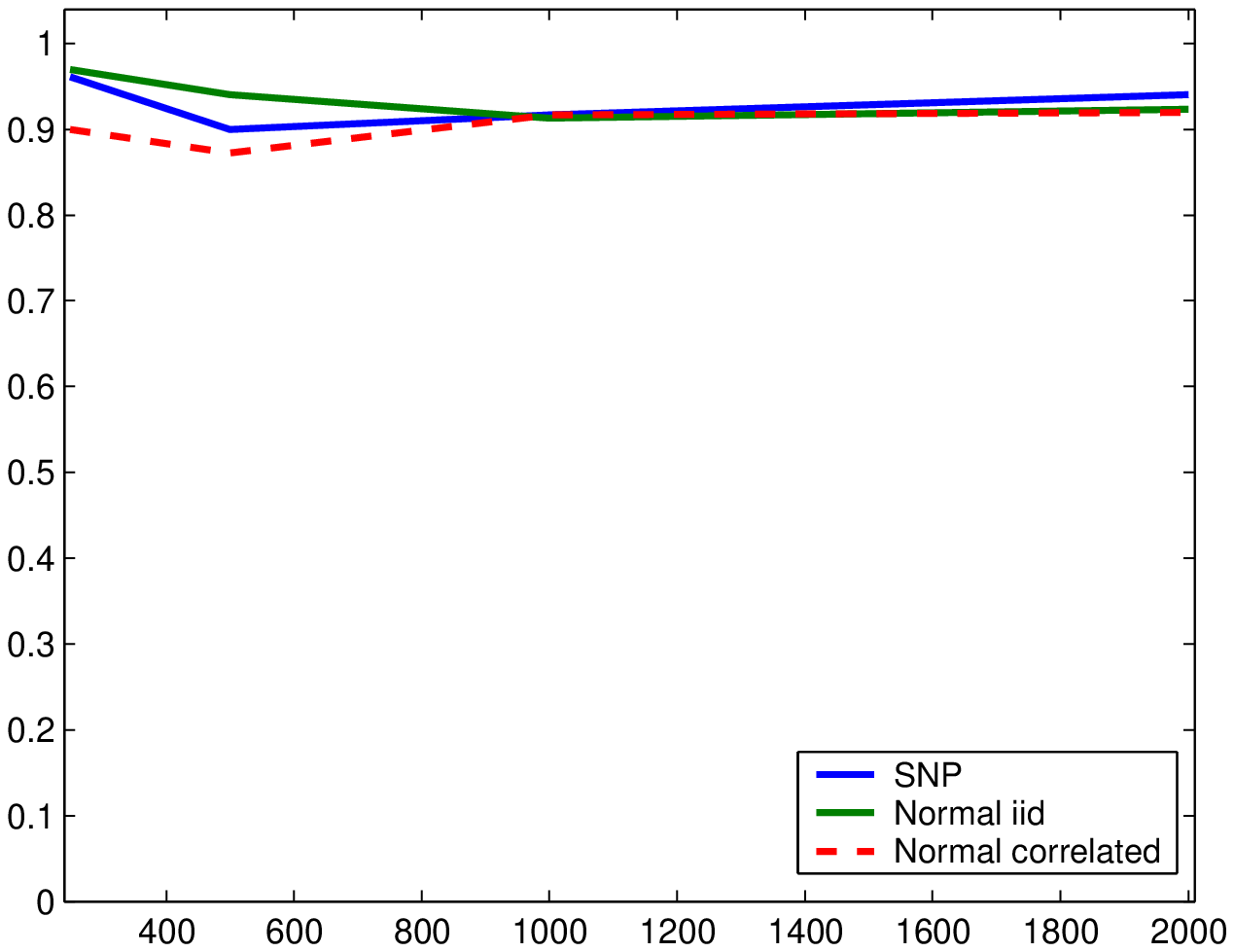}
\includegraphics[width=5.5cm]{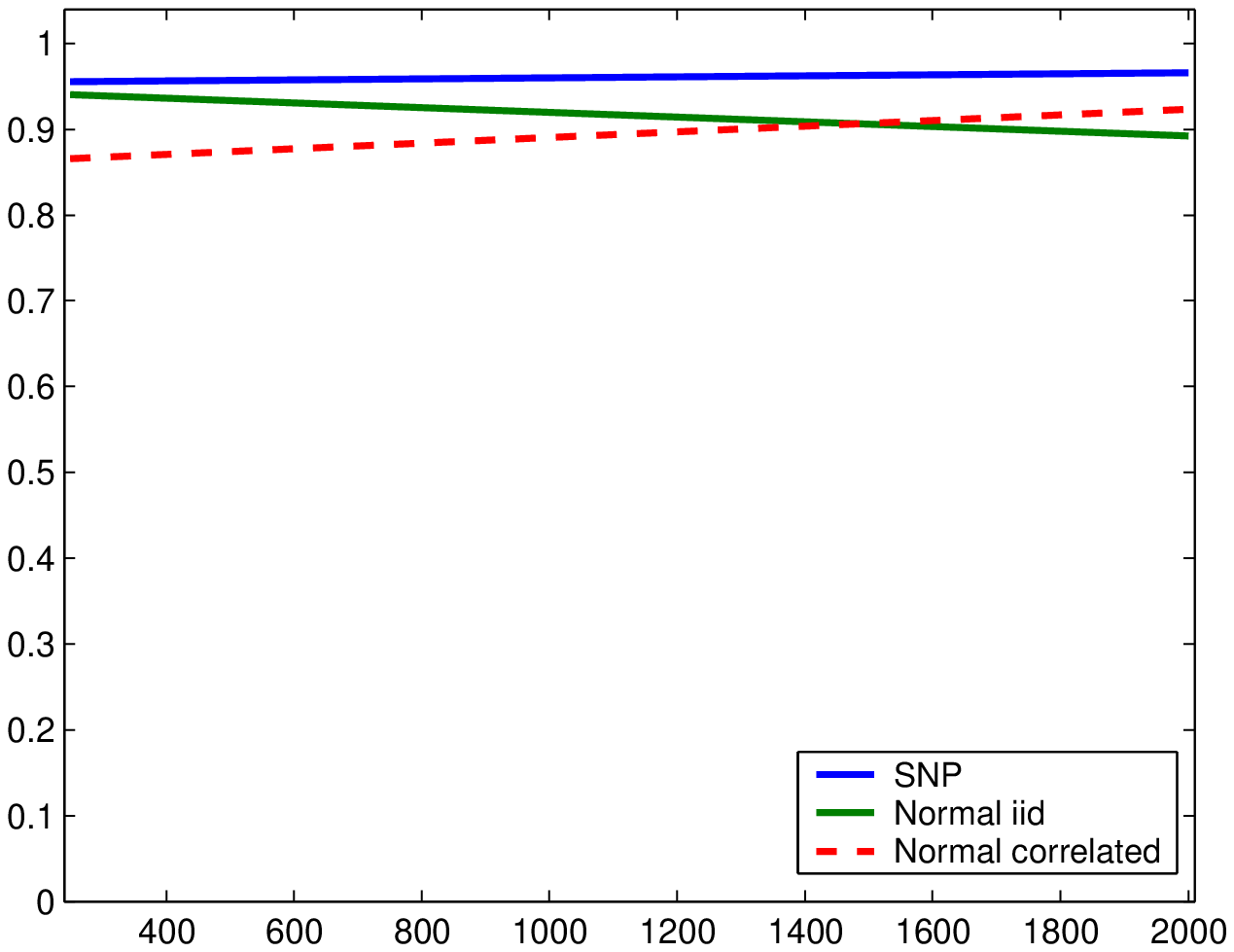}
\includegraphics[width=5.5cm]{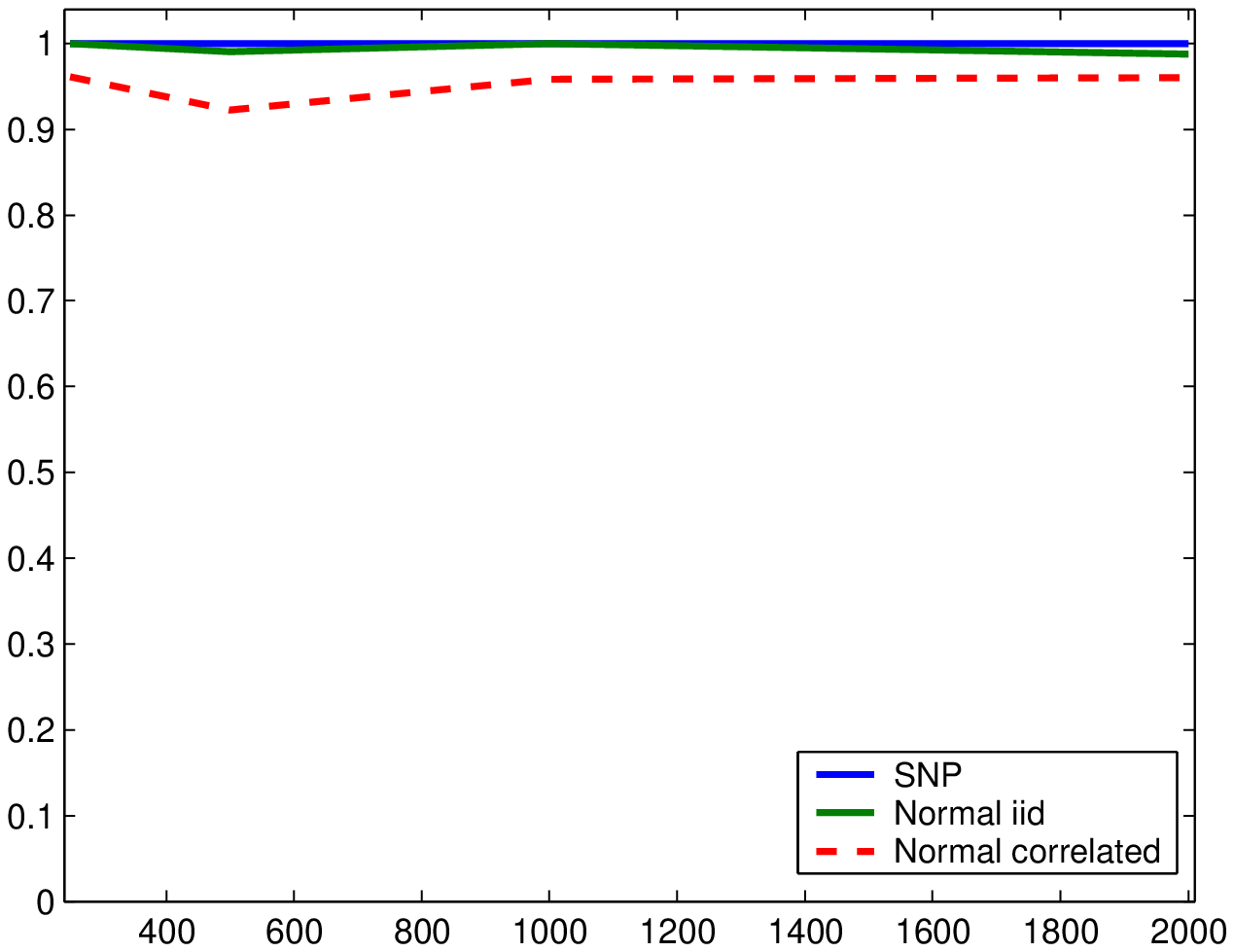}
\includegraphics[width=5.5cm]{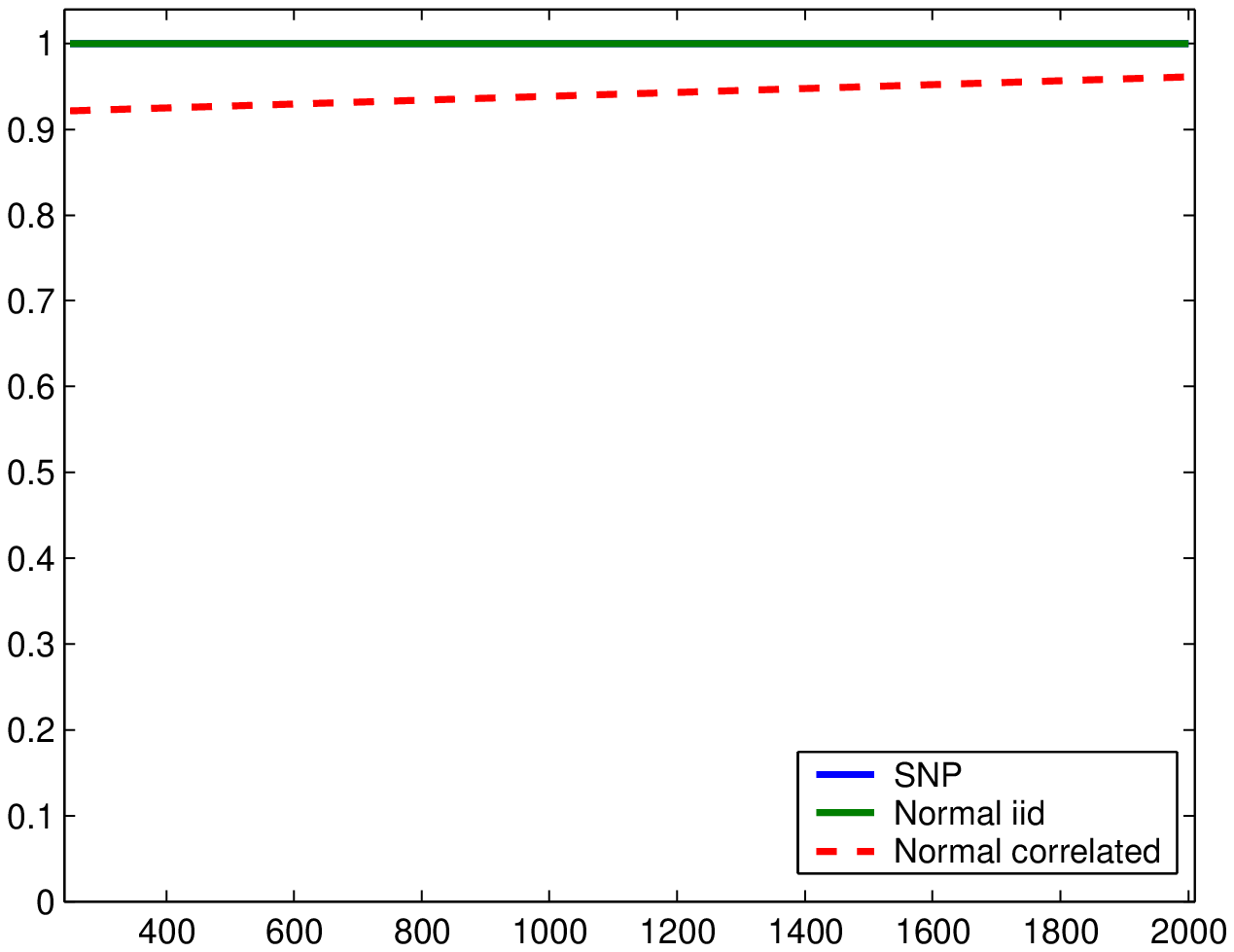}

\caption{The percentage of times $\hat I=I^*$(top row) and   $I^* \subseteq \hat I$(bottom row) vs number $M$ of predictors. Left column: $k^*=3, 2n=300$,  right column: $k^*=10, 2n=1000$.}
 \label{fig:eye_m}
 \end{figure}

In this section we consider the same simulation scenario as above, and we shift focus to   the quality of variable selection.  In Figure \ref{fig:eye_m} below we investigated  the sensitivity of the estimated probability of correct variable selection, $I^* = \hat I$, or correct inclusion, $I^* \subseteq \hat I$, as  we varied  $M = 250, 500, 1000, 2000$. In the interest of space, we only report below  the cases $k^* = 3, 2n = 300$ (left column) and $k^*=10, 2n = 1000$ (right column). 

\begin{figure}[ht]
\centering
\includegraphics[width=5.5cm]{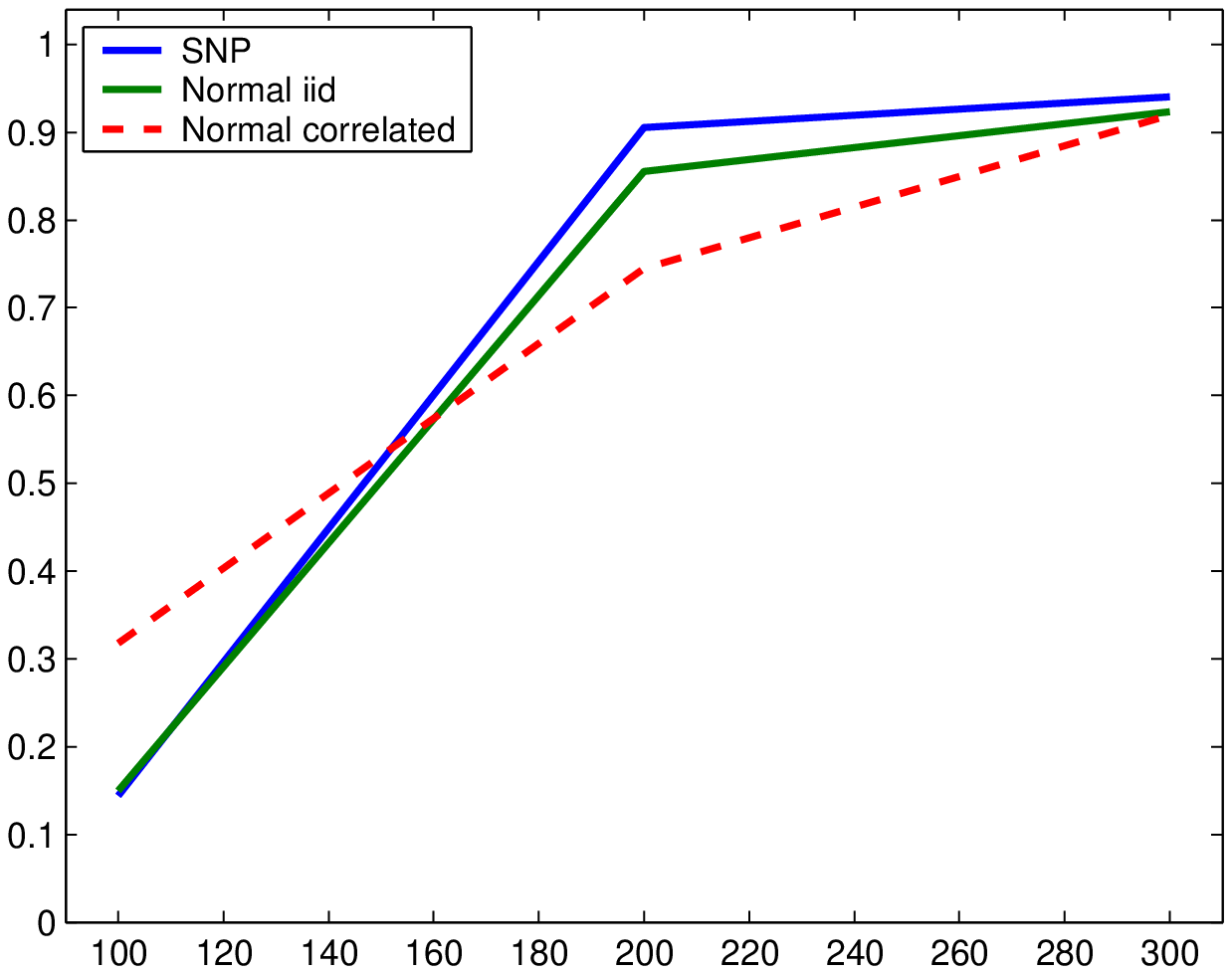}
\includegraphics[width=5.5cm]{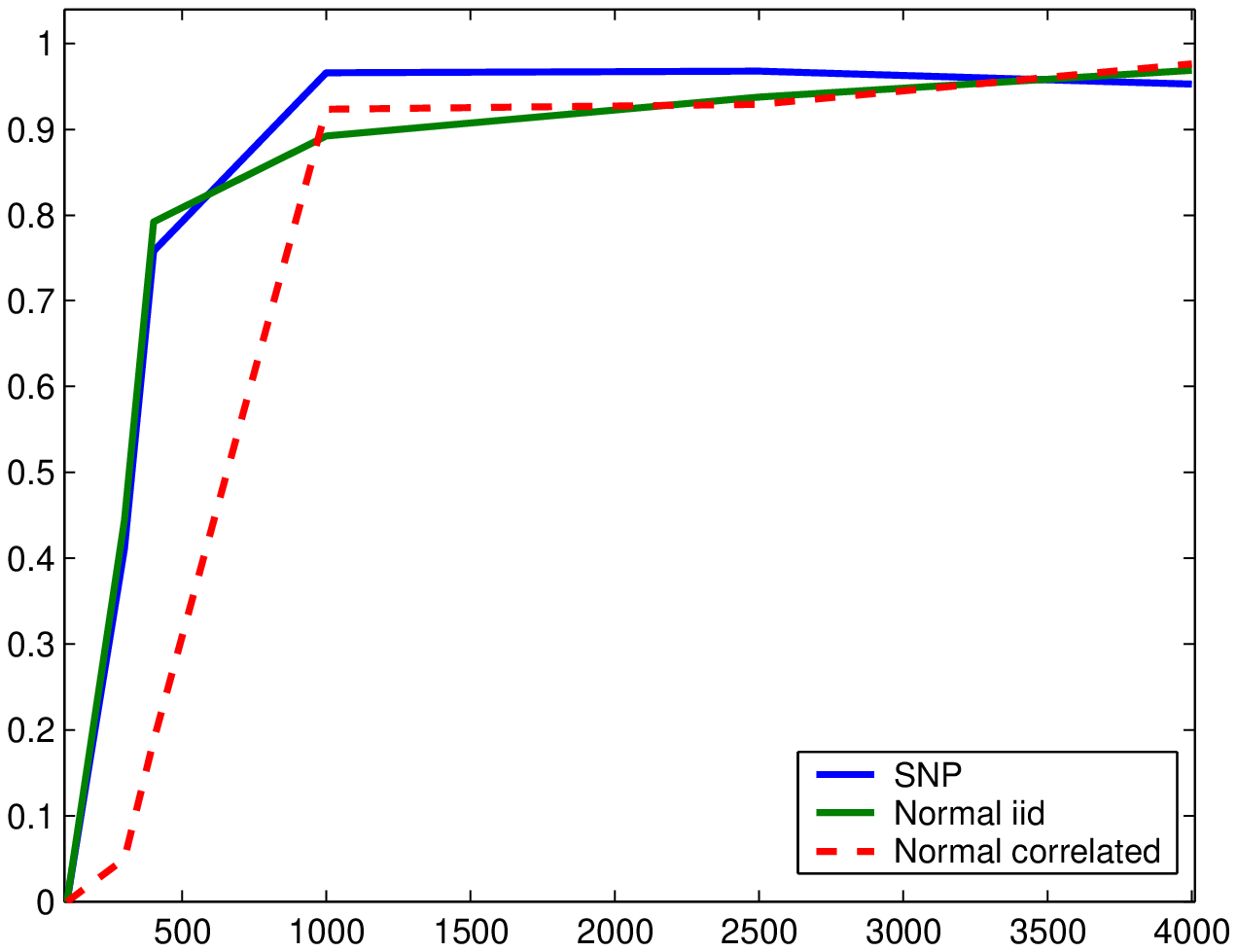}
\includegraphics[width=5.5cm]{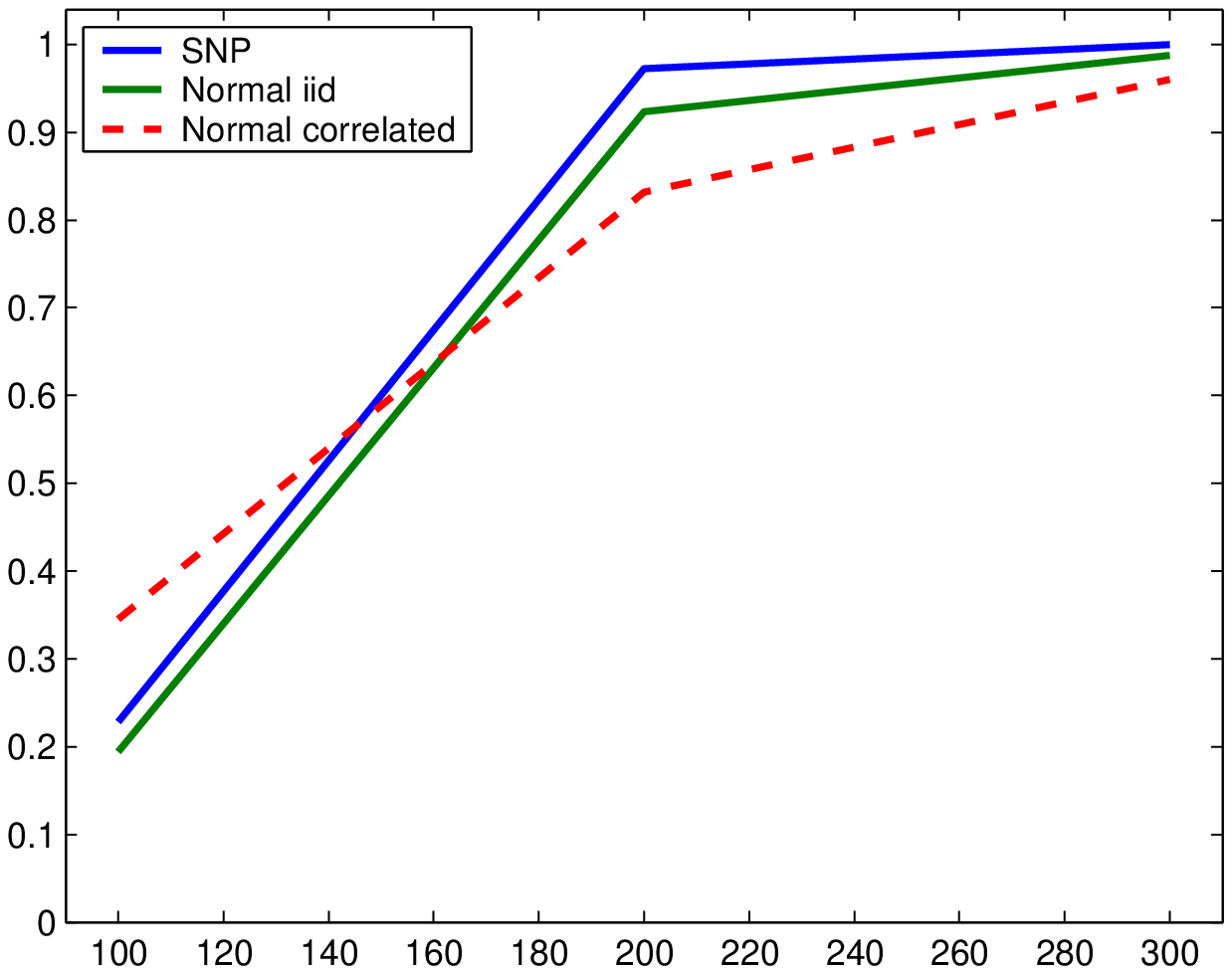}
\includegraphics[width=5.5cm]{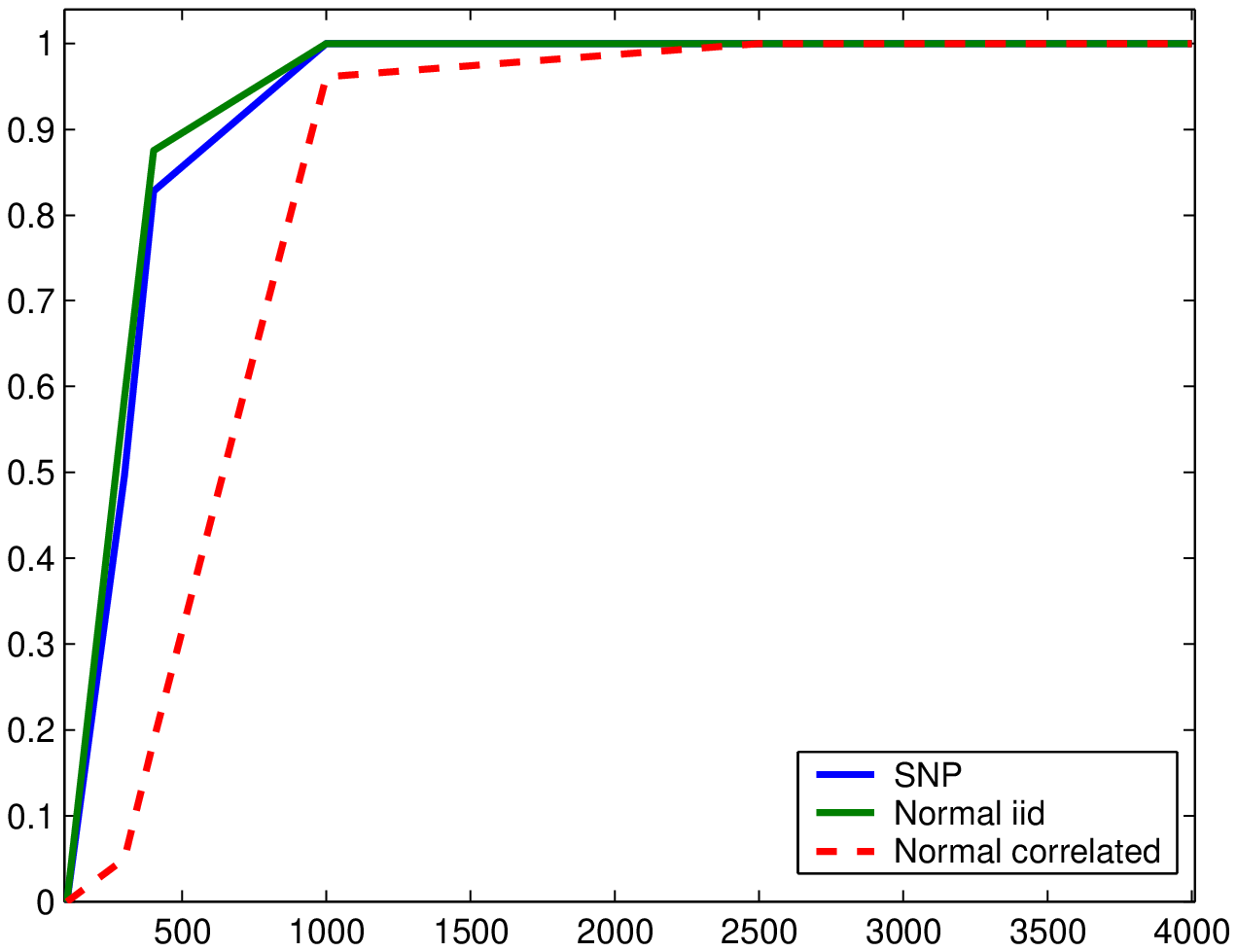}
\caption{The percentage of times $\hat I=I^*$(top row) and  $I^* \subseteq \hat I$(bottom row)  vs. number $2n$ of observations; $M = 2000$. Left column:  $k^* = 3$, right column: $k^*=10$.}
\label{fig:eye_n}
\end{figure}
We present, in Figure \ref{fig:eye_n} below, graphs of the percentage of times $I^* = \hat I$ and, respectively, $I^* \subseteq \hat I$,  for $M = 2000$, as we varied $n$, for $k^* = 3$ and $k^* = 10$. For ease of reference, we summarized in Table \ref{tab:samplsel} the sample sizes needed to identify the true model at least 90\% of the time. 
\begin{table}[ht]
\small
\begin{center}
\begin{tabular}{l c c c}
  &    
  SNP &NOR\_IID &NOR\_CORR\\
    \hline
$k^* = 3$ &200 &250 &300 \\
$k^* = 10$ &800 &1000 &1000\\
\hline
\end{tabular}
\vskip 1mm
\caption{Sample sizes needed for $\PP(\widehat{I} = I^*) \geq 0.90$; M = 2000.\label{tab:samplsel}
}
\end{center}
\end{table}

The results of this section support strongly the theoretical results of Section 4. Our method 
can identify the correct model with high probability. The accuracy of selection is influenced by the true model size $k^*$ and the dependence structure of the variables $X_j$, as shown in Figure \ref{fig:eye_n} and it is much less influenced  by an increase in the total number of the variables in the model, as shown in Figure \ref{fig:eye_m}.

\section{Conclusions}

\noindent We summarize our overall contributions  in this section. \\

\noindent 1. In this article we offered a  theoretical analysis of the quality of  model selection-type  estimators in case-control studies. Our focus has been on  optimizers of the $\ell_1$ regularized logistic likelihood.  We showed that these estimators, analyzed under the case-control sampling scheme have  model selection properties  that are similar to those of  the estimates analyzed under the prospective sampling scheme. In particular, we established the after model selection consistency of the odds ratio, and the consistency  of subset selection.  To the best of our knowledge, this is the first  such theoretical analysis conducted for this sampling scheme.\\

\noindent 2. We introduced a computationally efficient variable selection  and dimension reduction procedure that uses a generalization of the Bisection Method, the {\bf GBM}.   We used the {\bf GBM}
to find simultaneously $M$ tuning parameters, each yielding an estimator with exactly $k$ non-zero entries, $ 1 \leq k \leq M$. The final estimator is selected from the set of these $M$ candidates as the minimizer of a $p$-fold cross-validated log-likelihood to which we added a $BIC$-type penalty. This technique is general and can be used  in connection with other loss functions and sparsity inducing penalty terms. The full theoretical investigation of this promising method is the subject of future research.  All our simulation experiments indicate very good performance of this procedure in all the scenarios we considered. Moreover, our technique provides important computational savings over  grid search based methods.

\section*{Appendix}

\noindent{\bf Proof of Lemma \ref{samelik}.}
Let $(\widehat{\delta}, \widehat{\beta})$ and  $\widehat{q}$ be given by (\ref{unc}).  
We show that the maximum value of $\log L_{retros}(\delta, \beta, q) + pen(\beta)$, over all 
$({\delta}, {\beta}, {q})$ that satisfy the constraint (\ref{constraint}), is bounded above and below by $\log L_{retros}(\widehat{\delta}, \widehat{\beta}, \widehat{q}) + pen(\widehat{\beta})$. The bound from above is immediate, as a constraint maximum is always smaller or equal than the unconstrained maximum.  To show the bound from below, we only need  to verify that $(\widehat{\delta}, \widehat{\beta}, \widehat{q})$ given by (\ref{unc}) satisfy the constraint (\ref{constraint}). \\

\noindent   Let $G(\delta, \beta) = \log L_{pros}(\delta, \beta) + pen(\beta)$. If $(\widehat{\delta}, \widehat{\beta})$ are given by (\ref{unc}), and $pen({\beta})$ is not a function of $\delta$ then, in particular, we have : 
\[ 0 = \frac{\partial \log G(\widehat{\delta}, \widehat{\beta})}{\partial \delta}= \frac{\partial \log L_{pros}(\widehat{\delta}, \widehat{\beta})}{\partial \delta} = n - \sum_{i=1}^{n}\widehat{p}_1(x^0_i) - \sum_{i=1}^{n}\widehat{p}_1(x^1_i),\]
\noindent and also 
\[ n =  \sum_{i=1}^{n}\widehat{p}_0(x^0_i) + \sum_{i=1}^{n}\widehat{p}_0(x^1_i),\]
\noindent where $\widehat{p}_j$ denotes $p_j$ evaluated at $(\widehat{\delta}, \widehat{\beta})$. Recall that  the maximum likelihood estimator of $q$ is 
\[ \widehat{q}(x) = \frac{1}{2n}\left(\sum_{i=1}^{n}\delta_{x_{i}^{0}} (x)+ \sum_{i=1}^{n}\delta_{x_{i}^{1}}(x)\right),\]
\noindent where $\delta_a$ denotes the Dirac function.  Then, condition (\ref{constraint}) is satisfied by $(\widehat{\delta}, \widehat{\beta}), \widehat{q}$, since 
\[ \int \widehat{p}_j(x)\widehat{q}(x)dx = \frac{1}{2n}\left(  \sum_{i=1}^{n}\widehat{p}_j(x^0_i) + \sum_{i=1}^{n}\widehat{p}_j(x^1_i)\right) = \frac{1}{2} .\]

\noindent This concludes the proof of this Lemma. $\blacksquare$ \\

\bigskip

\noindent {\bf Proof of Theorem \ref{delta}.} 
\noindent We begin by introducing the notation used in the sequel. Let $\PP_n^0 =: \frac{1}{n}\sum_{i=1}^{n}\delta_{{X}_{i}^{0}}$ and $\PP_n^1=: \frac{1}{n} \sum_{i=1}^{n}\delta_{{X}_{i}^{1}}$ denote the empirical measures associated with the two samples. 
 For any function $g$ of generic argument $x$  we will use the  notation  $\PP_n^j g (x)=  \frac{1}{n}\sum_{i=1}^{n}g({ x}_i)$, for $j = 0, 1$.  Recall that we defined  the functions 
 \begin{eqnarray}   \ell_0(\theta) &= :&   \ell_0(\theta; x) \  = \ \log ( 1 + e^{\theta^{\prime}x}), \nonumber \\
\ell_1(\theta) &=: & \ell_1( \theta; x) \ = \ \log ( 1 + e^{\theta^{\prime}x}) - \theta^{\prime}x,\nonumber \end{eqnarray} 
\noindent and recall the notation $\theta = (\delta, \beta)$.  Then, by the definition of the estimator we have 
\begin{eqnarray}\label{defe}\frac{1}{2} \left (\PP_n^0\ell_0(\widehat{\theta}) + \PP_n^1\ell_1(\widehat{\theta})\right ) + 2r\sum_{j=1}^{M}|\widehat{\beta}_j| 
&\leq&  \frac{1}{2} \left (\PP_n^0\ell_0(\theta) +   \PP_n^1\ell_1(\theta)\right )\\
&&  \  \  \  \  \  \  
  + \ 2r\sum_{j=1}^{M}|{\beta}_j|, \nonumber
\end{eqnarray}
for all $\theta = (\delta, \beta)$. In particular, for $\theta = 0$, this shows that 
\[ \sum_{j=1}^{M}|\widehat{\beta}_j|\leq \frac{\log 2}{2r} \ \ \ \ \ \ \mbox{and} \  \ \ \ \ \ |\widehat{\delta}| \leq \frac{L\log 2}{2r},\]
\noindent where $L$ is a common bound on $X_{ij}^0, X_{ij}^1$ for all $i$ and $j$. 
Therefore, the estimator is effectively computed on the parameter set 
\begin{eqnarray}\label{C} \mathcal{C} = \left \{ \beta \in \RR^M, \delta \in \RR: \ \ \sum_{j=1}^{M}|{\beta}_j|\leq \frac{\log 2}{2r}, \  |{\delta}| \leq \frac{L\log 2}{2r} \right \},\end{eqnarray} 
\noindent and it is therefore enough to restrict our study to this set. 
\bigskip 

\noindent 

Recall that we defined $\Delta(\widehat{\theta}, \theta^*)$ in (\ref{deltadif}) as 
\[ \Delta(\widehat{\theta}, \theta^*) = \frac{1}{2} \PP^0 \left( \ell_0 (\widehat{\theta})  - \ell_0(\theta^*) \right)    +   
\frac{1}{2}\PP^1 \left(    \ell_1  (\widehat{\theta}) -   \ell_1(\theta^*)  \right).\]

\noindent   Note further that since  (\ref{defe})  holds for all $\theta$ it holds in particular for for $\theta = \theta^*$.  Then, by adding $\Delta(\widehat{\theta}, \theta^*) + r\sum_{j=1}^{M}|\widehat{\beta}_j - \beta_j^*|$  to both sides of   (\ref{defe}) and re-arranging terms  we obtain
\begin{eqnarray}\label{int}&& r\sum_{j=1}^{M}|\widehat{\beta}_j - \beta_j^*| +  \Delta(\widehat{\theta}, \theta^*)    \leq   \frac{1}{2}(\PP_n^0 - \PP^0) \left( \ell_0(\theta^*) - \ell_0  (\widehat{\theta}) \right) \\
&&\hspace{4.5cm} + \
  \frac{1}{2}(\PP_n^1 - \PP^1)\left( \ell_1(\theta^*) - \ell_1 (\widehat{\theta}) \right) \nonumber \\
  &&\hspace{4.5cm}  + \ r\sum_{j=1}^{M}|\widehat{\beta}_j - \beta_j^*| + 2r\sum_{j=1}^{M}|\beta_j^*|-
 2 r\sum_{j=1}^{M}|\widehat{\beta}_j|. \nonumber \end{eqnarray}

\noindent  Let $\epsilon > 0$ be a quantity that will be made precise below. Then, for  $\theta \in \mathcal{C}$ given by  (\ref{C}) above we define  \begin{eqnarray}  \mathcal{L}_n^0 &=&  \sup_{\theta \in \mathcal{C}} \frac {|(\PP_n^0 - \PP^0) ( l_0(\theta^*) -  l_0({\theta}) )| }{|\beta - \beta^*|_1 + \frac{|\delta - \delta^*|}{\log n} +  \epsilon}, \nonumber \\
 \mathcal{L}_n^1 &=&  \sup_{\theta \in \mathcal{C}} \frac {|(\PP_n^1 - \PP^1) ( l_1(\theta^*) -  l_1({\theta}) )| }{|\beta - \beta^*|_1 + \frac{|{\delta} - \delta^*|}{\log n} +  \epsilon}, \nonumber
 \end{eqnarray}
\noindent where $|v|_1 = \sum_{j=1}^{d}|v_j|$ denotes the $\ell_1$ norm of a generic vector $v \in \RR^d$, for some $d$.  Define the events \begin{eqnarray}\label{ee} \mathcal{E}_0 =\left\{ \mathcal{L}_n^0 \leq r\},  \  \mathcal{E}_1 = \{\mathcal{L}_n^1 \leq r \right\}.\end{eqnarray}

\noindent Then, on $ \mathcal{E}_0 \cap \mathcal{E}_1$ display (\ref{int}) above yields 
\begin{eqnarray}\label{unu}
&& r\sum_{j=1}^{M}|\widehat{\beta}_j - \beta_j^*| +  \Delta(\widehat{\theta}, \theta^*) \\
&& \hspace{0.3cm}\leq   \ 2r\sum_{j=1}^{M}|\widehat{\beta}_j - \beta_j^*| + 2r\sum_{j=1}^{M}|\beta_j^*|-
 2 r\sum_{j=1}^{M}|\widehat{\beta}_j| \nonumber \\
 && \hspace{0.3cm} + \ r\frac{|\widehat{\delta} - \delta^*|}{\log n} +  r\epsilon. \nonumber 
\end{eqnarray}

\medskip 

\noindent We will argue below  that: \\

\noindent {\it Fact 1.}  $\Delta(\widehat{\theta}, \theta^*) \geq  0$. \\

\noindent {\it Fact 2.} $ \PP(\mathcal{E}_0^c) \rightarrow 0$ and $\PP (\mathcal{E}_1^c) \rightarrow 0$. \\

\noindent  Assume that {\it Fact 1} and {\it Fact 2} hold. Then, if {\it Fact 1} holds, both terms in the left hand side of display (\ref{unu})  are positive.  Thus, in particular,  display (\ref{unu}) yields on the event $ \mathcal{E}_0 \cap \mathcal{E}_1$ that 
\begin{eqnarray} 
&& \Delta(\widehat{\theta}, \theta^*) \leq   \ 2r\sum_{j=1}^{M}|\widehat{\beta}_j| +  2r\sum_{j=1}^{M}|\beta_j^*| + 2r\sum_{j=1}^{M}|\beta_j^*|-
 2 r\sum_{j=1}^{M}|\widehat{\beta}_j| \nonumber \\
 && \hspace{2cm}+ \ r\frac{|\widehat{\delta} - \delta^*|}{\log n} +  r\epsilon \nonumber \\
 && \hspace{1.9cm} \leq 4r\sum_{j \in I^*}|\beta_j^*| +  r\frac{|\widehat{\delta} - \delta^*|}{\log n} +  r\epsilon.  \nonumber \end{eqnarray}

\noindent Recall now that we have assumed that $\max_{j \in I^*}|\beta_j^*| \leq B, |\delta^*|\leq B$, for some positive constant $B$ and that $\widehat{\delta} \in \mathcal{C}$, and so $|\widehat{\delta}| \leq \frac{L\log 2}{2r}$. Thus, on the event $\mathcal{E}_0 \cap \mathcal{E}_1$,  we have 
\begin{equation}
 \Delta(\widehat{\theta}, \theta^*) \leq 4rk^*B + \frac{L\log 2}{2\log n} +  r\frac{B}{\log n} + r\epsilon. \nonumber
\end{equation}
\noindent Then, for our choice of $r$, under the assumption that $rk^* \rightarrow 0$ and for the choice of $\epsilon$ given below in (\ref{eps}), the right hand side of the above display converges to zero with $n$. Therefore, for any $\alpha > 0$ we have 
the desired result: 
\begin{equation}\label{conv0}
\PP(  \Delta(\widehat{\theta}, \theta^*) > \alpha ) \leq \PP(\mathcal{E}_0^c) +  \PP (\mathcal{E}_1^c) \longrightarrow 0,
\end{equation}
\noindent provided {\it Fact 2} holds. We argue in what follows that the two facts hold. \\

\noindent {\it Proof of Fact 1.} Recall the definition (\ref{crux}) of $f_0(x)$, $f_1(x)$, $q(x)$  and $p_1(x)$, and notice that we have $f_0(x) + f_1(x) = 2q(x)$.  Let $\bar{\theta}^{\prime}x$ be an intermediate point between $ \theta^{*\prime}x$ and 
 $\widehat{\theta}^{\prime}x$. Then, a second order Taylor expansion gives:
 \begin{eqnarray}\label{deltalong}
 \Delta(\widehat{\theta}, \theta^*) & = & \int \left( \log (1 + \exp(\widehat{\theta}^{\prime}x)) - \log (1 + \exp({\theta}^{*\prime}x)) \right)q(x)dx  \\
 && -  \int (\widehat{\theta}^{\prime}x- {\theta}^{*\prime}x)p_1(x)q(x)dx  \nonumber \\
 & = & \int (\widehat{\theta}^{\prime}x - {\theta}^{*\prime}x)p_1(x)q(x)dx  \nonumber \\
&&  + \  \frac{1}{2}\int\frac{\exp( \bar{\theta}^{\prime}x ) }{ (1 + \exp(  \bar{\theta}^{\prime}x ) )^2 } (\widehat{\theta}^{\prime}x - {\theta}^{*\prime}x)^2q(x)dx  \nonumber \\
 &&   - \int (\widehat{\theta}^{\prime}x - {\theta}^{*\prime}x)p_1(x)q(x)dx  \nonumber \\
 &=&  \frac{1}{4}\int\frac{\exp( \bar{\theta}^{\prime}x ) }{ (1 + \exp(  \bar{\theta}^{\prime}x ) )^2 } (\widehat{\theta}^{\prime}x - {\theta}^{*\prime}x)^2f_0(x)dx  \nonumber \\
 && + \   \frac{1}{4}\int\frac{\exp( \bar{\theta}^{\prime}x ) }{ (1 + \exp(  \bar{\theta}^{\prime}x) )^2 } (\widehat{\theta}^{\prime}x - {\theta}^{*\prime}x)^2f_1(x)dx,    \nonumber 
  \end{eqnarray}
\noindent which shows that the left hand side is positive. \\

\noindent {\it Proof of Fact 2.} Define the re-scaled empirical processes 
\begin{eqnarray} 
\GG_n^0 = \sup_{\theta \in \mathcal{C}} \frac {|(\PP_n^0 - \PP^0) ( l_0(\theta^*) -  l_0({\theta}) )| }{|\beta - \beta^*|_1 + |\delta - \delta^*| +  \epsilon} \  \ \mbox{and}, \nonumber \\
 \GG_n^1 = \sup_{\theta \in \mathcal{C}} \frac {|(\PP_n^1 - \PP^1) ( l_1(\theta^*) -  l_1({\theta}) )| }{|\beta - \beta^*|_1 + |{\delta} - \delta^*|+  \epsilon}, \nonumber
 \end{eqnarray}
\noindent and notice that 

\begin{equation}\label{almost}   \PP(\mathcal{E}_0^c) \leq  \PP\left (\GG_n^0   > \frac{r}{\log n}\right )  \ \  \ \mbox{and}  \ \ \ \PP (\mathcal{E}_1^c) \leq \PP\left (\GG_n^1 > \frac{r}{\log n} \right ).\end{equation}
\noindent  The proof of {\it Fact 2}  therefore relies on the control of $\GG_n^0$ and $\GG_n^1$.  For this, we use  the bounded difference inequality, see e.g. Theorem 2.2, page 8 in \cite{luc}. To apply it we need to evaluate by how much 
 ${\GG}_n^0$ and  $  {\GG}_n^1$  change  if we change  the $i$-th variable $X_i^0$,  $X_i^1$, respectively, while keeping the others fixed.  Recall that $\PP_n^0 = \frac{1}{n}\sum_{i=1}^{n}\delta_{X_i^0}$ is the empirical measure putting mass $1/n$ at each observation $X_i$. Let $\PP^{0\prime}_n$ be the empirical measure $\frac{1}{n}\left(\sum_{i=1, i \neq l}^{n-1}\delta_{X_i^0} + \delta_{X_{l}^{0\prime}}\right)$ corresponding to changing the pair $X_{l}^0$ to $X_{l}^{0\prime}$.
Then 
\begin{eqnarray}  &&\frac {(\PP_n^0 - \PP^0) ( l_0(\theta^*) -  l_0({\theta}) ) }{|\beta - \beta^*|_1 + |\delta - \delta^*| +  \epsilon} - \frac {(\PP^{0\prime}_n - \PP^0) ( l_0(\theta^*) -  l_0({\theta}) ) }{|\beta - \beta^*|_1 +  |\delta - \delta^*| + \epsilon}  \nonumber \\
& & \hspace{0.2cm}=  \frac{1}{n} \frac{ l_0(\theta^*;  X_l^0) -  l_0({\theta};  X_l^0) -  l_0(\theta^*; X_l^{0\prime})  +  l_0({\theta};  X_l^{0\prime})  }{   |\beta - \beta^*|_1 + |\delta - \delta^*| + \epsilon} \nonumber \\
& & \hspace{0.2cm} \leq \frac{4L}{n} \frac{  |\beta - \beta^*|_1 + |\delta - \delta^*|  } { |\beta - \beta^*|_1 +  |\delta - \delta^*| + \epsilon}  \leq \frac{4L}{n},
\end{eqnarray}
\noindent where the inequality follows immediately by a first order Taylor expansion and the assumption that all $X$ variables are bounded by $L >1$.  The calculations involving 
${\GG}^1_n$ are identical, and yield the bound $\frac{8L}{n}$. Therefore, we can apply the bounded difference inequality to obtain that 
\begin{eqnarray}\label{expb} \PP^0 ( {\GG}_n^0 - \EE^0 {\GG}^0_n  \geq  u) &\leq &\exp{-\frac{nu^2}{8L^2}} , \\
 \PP^1 ( {\GG}_n^1 - \EE^1 {\GG}^1_n  \geq  u) &\leq &\exp{-\frac{nu^2}{32L^2}} \nonumber 
.\end{eqnarray}

\noindent We will use Lemma 3 in \cite{mart} to obtain  bounds on $\EE^0\GG_n^0$ and $\EE^1\GG_n^1$. We re-state a version of it here,  for ease of reference. \\

\noindent {\it  Let $J_n$ be an integer such that $2^{J_{n}} \geq n$ and $0 < \epsilon \leq  \frac{\log 2}{r}$ Then, if the functions $l_0$ and $l_1$  defined above are  Lipschitz in $\theta^{\prime}x$  and the components of $x$ are  bounded  by $L$, with probability one,  then both
$\EE^0 \GG_n^0,  \ \EE^1\GG_n^1$ are bounded by  $ C_1\sqrt{\frac{2\log 2 (M \vee n)}{n}} +  C_2\frac{J_n}{  2(M \vee  n)^2}$, where $C_1, C_2$ are positive constants depending on the respective Lipschitz constants and $L$.} \\

\noindent Notice that  $l_0$ and $l_1$ are Lipschitz   in $t = \theta^{\prime}x$, with respective constants 1 and 2. Also, inspection of the chaining argument used in the proof of this lemma shows that we can take 
$J_n = (M \vee n)$ and \begin{equation}\label{eps}\epsilon= \frac{\log 2}{2^{(M \vee n) + 1}}\times \frac{1}{r},\end{equation}
\noindent for $r$ given in (\ref{er}), which we also recall here \[ r =  \log n  \left(6L \sqrt{\frac{2\log 2 (M \vee n)}{n}} + \frac{1}{  4(M \vee  n)}  +  4L\sqrt{\frac{2\log\frac{1}{\delta}}{n}}  \right). \]

\noindent Then,  making the constants precise in the lemma above and   taking  \begin{equation}\label{u} u = 4L\sqrt{\frac{2\log\frac{1}{\delta}}{n}},\end{equation}
\noindent display (\ref{expb}) yields
\[ \PP^0 \left( {\GG}_n^0  \geq  \frac{r}{\log n}\right) \leq \delta, \ \ 
 \PP^1 \left ( {\GG}_n^1 \geq  \frac{r}{\log n} \right ) \leq \delta, \]
\noindent for any $\delta >0$, in particular for any $\delta = \delta_n \rightarrow 0$.
This display in combination with (\ref{almost}) above gives the desired result. This completes the proof of this theorem. \ $\blacksquare$

\bigskip 

\noindent {\bf Proof of Corollary \ref{firstsup}.} By definition 
 \begin{eqnarray}
 \Delta(\widehat{\theta}, \theta^*) & = & \int \left( \log (1 + \exp(\widehat{\theta}^{\prime}x)) - \log (1 + \exp({\theta}^{*\prime}x)) \right)q(x)dx \nonumber \\
 && -  \int (\widehat{\theta}^{\prime}x - {\theta}^{*\prime}x)p_1(x)q(x)dx  \nonumber \\
 &=& \int (\widehat{\theta}^{\prime}x - {\theta}^{*\prime}x)\left( \frac{\exp( \bar{\theta}^{\prime}x)}{ 1  + \exp(\bar{\theta}^{\prime}x)} - \frac{\exp( {\theta}^{*\prime}x)}{ 1  + \exp({\theta}^{*\prime}x)} \right)q(x)dx, \nonumber
 \end{eqnarray}
\noindent where the last line follows via a first order Taylor expansion. Simple algebraic manipulations  show  that  if $\sup_{x}|\widehat{\theta}^{\prime}{x} - {\theta}^{*\prime}{x}| \geq \gamma$, for any $\gamma >0$, then $\Delta(\widehat{\theta}, \theta^*) >0$. Therefore, there exits $\alpha_{\gamma}$ such that  $\Delta(\widehat{\theta}, \theta^*) \geq \alpha_{\gamma}$. Invoking Theorem \ref{delta} above  we therefore obtain 
\begin{equation}\label{boundgood}
\PP\left(\sup_{x}|\widehat{\theta}^{\prime}{x} - {\theta}^{*\prime}{x}| \geq \gamma\right) \leq \PP\left (  \Delta(\widehat{\theta}, \theta^*) > \alpha_{\gamma}\right) \longrightarrow 0,
\end{equation}
\noindent which is the desired result. $\blacksquare$ \\

\bigskip

\noindent {\bf Proof of Theorem \ref{rates}}. Recall that in  (\ref{deltalong}) above we showed that 
 \begin{eqnarray}
 \Delta(\widehat{\theta}, \theta^*) 
  &=&  \frac{1}{4}\int\frac{\exp( \bar{\theta}^{\prime}x) }{ (1 + \exp(  \bar{\theta}^{\prime}x ) )^2 } (\widehat{\theta}^{\prime}x - {\theta}^{*\prime}x)^2f_0(x)dx  \nonumber \\
 && + \   \frac{1}{4}\int\frac{\exp( \bar{\theta}^{\prime}x ) }{ (1 + \exp(  \bar{\theta}^{\prime}x ) )^2 } (\widehat{\theta}^{\prime}x - {\theta}^{*\prime}x)^2f_1(x)dx,    \nonumber 
  \end{eqnarray}
with  $\bar{\theta}^{\prime}x$ being an intermediate point between $ \theta^{*\prime}x$ and $\widehat{\theta}^{\prime}x$.
\bigskip 

\noindent Let $\gamma > 0$ be arbitrarily close to zero, fixed.  Let $A_{\gamma} = \{ \sup_{x}|\widehat{\theta}^{\prime}{x} - {\theta}^{*\prime}{x}| \leq \gamma \}$. By (\ref{boundgood}) above  we have $\PP(A_{\gamma}) \rightarrow 1$. On $A_{\gamma}$ we have ${\theta}^{*\prime}{x} \geq  \bar{\theta}^{\prime}x - \gamma$, 
for all $x$,  and therefore 
\begin{eqnarray}
\frac{\exp( \bar{\theta}^{\prime}x ) }{ (1 + \exp(  \bar{\theta}^{\prime}x) )^2 }  &=& \exp(\bar{\theta}^{\prime}x - {\theta}^{*\prime}x)  \left (\frac{1 + \exp({\theta}^{*\prime}x)}{   1 + \exp(\bar{\theta}^{\prime}x )}\right)^2p_0(x)p_1(x) \nonumber \\
& \geq & \frac{ \exp(\bar{\theta}^{\prime}x - {\theta}^{*\prime}x)   }  { \exp(2\gamma) } \nonumber \\ 
&\geq & \exp(-3\gamma)p_0(x)p_1(x) =: wp_0(x)p_1(x), \nonumber
\end{eqnarray}
for all $x$ and with $w$ arbitrarily close to one.  Recall that $f(x) = \pi f_0(x) + (1 - \pi) f_1(x)$, for $\PP(Y = 1) = \pi$. Then, on the set $A_{\gamma}$ we have 
 \begin{eqnarray}\label{cucu}
 \Delta(\widehat{\theta}, \theta^*)  &\geq&  \frac{w}{4}\int
    p_0(x)p_1(x)(\widehat{\theta}^{\prime}x - {\theta}^{*\prime}x)^2(f_1(x) + f_0(x))dx,    \\
    & \geq&  \frac{w}{4}\int
    p_0(x)p_1(x)(\widehat{\theta}^{\prime}x - {\theta}^{*\prime}x)^2f(x)dx \nonumber \\
  &\geq&  wb\left(\sum_{j \in I^*}(\widehat{\beta}_j - \beta_j^*)^2 + (\widehat{\delta} - \delta^*)^2\right),\nonumber
  \end{eqnarray}
\noindent where the last inequality follows from {\it Condition H}. Then, adding and subtracting $r|\widehat{\delta} -\delta^*|$ to both sides in  (\ref{unu}) and using  (\ref{cucu})  above we obtain 
\begin{eqnarray}\label{intdoi} 
&&
r|\widehat{\delta} - \delta^*| + r\sum_{j=1}^{M}|\widehat{\beta}_j - \beta_j^*| +    wb\sum_{j \in I^*}(\widehat{\beta}_j - \beta_j^*)^2 + wb(\widehat{\delta} - \delta^*)^2 
\nonumber \\
  && \leq  \ 2r\sum_{j=1}^{M}|\widehat{\beta}_j - \beta_j^*| + 2r\sum_{j=1}^{M}|\beta_j^*|-
 2 r\sum_{j=1}^{M}|\widehat{\beta}_j| + r(1 + \frac{1}{\log n})|\widehat{\delta} - \delta^*| + r\epsilon \nonumber  \\
 && \leq 4r\sum_{j \in I^*}|\widehat{\beta}_j - \beta_j^*| + r(1 + \frac{1}{\log n})|\widehat{\delta} - \delta^*| + r\epsilon \nonumber  \\
  && \leq \frac{6}{bw}r^2k^* +  wb\sum_{j \in I^*}(\widehat{\beta}_j - \beta_j^*)^2 
  + wb(\widehat{\delta} - \delta^*)^2,\nonumber 
  \end{eqnarray}

where we obtained the last line by using the  Cauchy-Schwarz inequality  followed by an inequality of the type $2uv \leq au^2 + v^2/a$, for any $a > 1$. For clarity of exposition, we absorbed the term $r\epsilon$, which is of order $1/2^{M \vee n}$, into the first term,  which is much larger. Therefore
\[ r|\widehat{\delta} - \delta^*| + r\sum_{j=1}^{M}|\widehat{\beta}_j - \beta_j^*| \leq   \frac{6}{bw}r^2k^*,\]
which implies 
\begin{eqnarray}
|\widehat{\delta} - \delta^*| &\leq&   \frac{6}{bw}rk^*, \nonumber 
\\ 
\sum_{j=1}^{M}|\widehat{\beta}_j - \beta_j^*| &\leq& \frac{6}{bw}rk^*, \nonumber 
\end{eqnarray}
on the set $A_{\gamma}$, with $\PP(A_{\gamma}) \rightarrow 1$, which is the desired result. $\blacksquare$ \\

\noindent{\bf Proof of Theorem \ref{sel}}.
Since
$ \PP( I^* = \hat{I}) \geq 1 - \PP(I^* \not \subseteq \hat{I}) - \PP(\widehat{I} \not \subseteq I^*) $, it is enough to control the two probabilities in the right hand side of this inequality separately. \\

\noindent {\it Control of  $\PP(I^* \not \subseteq \hat{I})$.} \\

\noindent Recall that we denoted the cardinality of $I^*$ by $k^*$. Then, by the definition of the sets $\widehat{I}$ and $I^*$ we have 
 \begin{eqnarray}\label{lower2}
\PP(I^* \not \subseteq \hat{I}) &\leq& \PP\left (k \notin \hat{I} \ {\mbox for \ some } \ k \in I^*\right) \nonumber \\
& \leq & k^* \max_{k \in I^*} \ \PP \left (\widehat \beta_{k} = 0 \ \mbox{and} \ \beta^*_{k} \neq 0\right)\nonumber  \\
& \leq & M \max_{k \in I^*} \ \PP \left (\widehat \beta_{k} = 0 \ \mbox{and} \ \beta^*_{k} \neq 0\right),\nonumber  
\end{eqnarray}
as we always have $k^* \leq M$. Recall that 
\noindent
\[ \log L_{pros}(\theta) \nonumber \\
 =   \sum_{i=1}^{n}\log ( 1 + e^{\theta^{\prime}{ X}_i^0})  -\sum_{i=1}^{n}( \theta^{\prime}{ X}_i^1)+ \sum_{i=1}^{n}\log( 1 + e^{\theta^{\prime}{X}_i^1} ),\]
and let $L_n(\theta) =  \log L_{pros}(\theta)$.  By standard results in convex analysis, if $\widehat{\beta}_k = 0$ is a component of the solution $\widehat{\theta} = (\widehat{\delta}, \widehat{\beta})$ then
\[ \left|\frac{\partial L_n(\widehat{\theta})}{\partial \beta_k} \right|  < 2r.\]

\noindent Let $k$ be arbitray, fixed.  Define \begin{eqnarray}\label{useful}T_n(\widehat{\theta}) =:T_n &= & \frac{\partial L_n(\widehat{\theta})}{\partial \beta_k}  
-  \frac{\partial L_n({\theta^*})}{\partial \beta_k} \\
S_n(\widehat{\theta}) =:S_n&=&  \sum_{j = 0}^{M} (\widehat{\theta}_j - \theta_j^*)\left(\frac{1}{2n}\sum_{i=1}^{2n}X_{ij}X_{ik}\right). \nonumber \end{eqnarray}

\medskip 

\noindent Recall that by convention $X_{i0} = 1$, for all $i$, and for compactness of notation we will write $\wh \theta_0 = \wh \delta$ and $\theta_{0}^{*} = \delta$. \\

\noindent Assuming without loss of generality  that the data has been scaled such that  $\frac{1}{2n} \sum_{i=1}^{n}X_{ik}^2 = 1$ for all $k$,  we therefore obtain,  for every $k \in I^*$, that 
\begin{align}\label{bigi}
&
\PP \left (\widehat \beta_{k} = 0 \ \mbox{and} \ \beta^*_{k} \neq 0\right) \nonumber \\
&\quad \leq \PP\left(\left| T_n - S_n + 	S_n
 +   \frac{\partial L_n({\theta^*})}{\partial \beta_k} 
 \right|
 \leq 2r; \ \ \beta_k^* \neq 0\right) \nonumber \\
&\quad \leq \PP\left( |S_n| - |S_n - T_n| - \left|  \frac{\partial L_n({\theta^*})}{\partial \beta_k}    \right|
 \leq 2r; \ \ \ \beta_k^* \neq 0 \right) \nonumber \\
 &\quad  \leq   \PP \left ( |\beta_k^*| \,{-}\, \Bigg|\sum_{j \neq k}(\widehat{\theta}_j \,{-}\, \theta_j^*)
 \frac{1}{2n}\sum_{i=1}^{2n}X_{ij}X_{ik}\Bigg|
  \,{-}\, |S_n - T_n| \,{-}\, \left| \frac{\partial L_n({\theta^*})}{\partial \beta_k}    \right|
 \leq   2r \right).\nonumber 
\end{align}

\noindent Therefore
\begin{eqnarray}\label{later}
&& \PP(I^* \not \subseteq \hat{I}) \leq M\max_{k \in I^*} \  \PP \left ( \left|   \frac{\partial L_n({\theta^*})}{\partial \beta_k}    \right| \geq   \frac{|\beta_k^*| - 2r}{3}  \right)  \\
&& \hspace{2cm}+ \  M\max_{k \in I^*}  \PP\left ( \sum_{j = 0}^{M} |\widehat{\theta}_j - \theta_j^*|
\Bigg|\frac{1}{2n}\sum_{i=1}^{2n}X_{ij}X_{ik}\Bigg| >   \frac{|\beta_k^*| - 2r}{3}   \right) \nonumber \\
&& \hspace{2cm} + \ M \max_{k \in I^*} \PP\left ( |S_n - T_n| \geq \frac{|\beta_k^*| - 2r}{3} \right) \nonumber \\
&& \hspace{1.8cm}  = (I) + (II) + (III). \nonumber
\end{eqnarray}

\noindent In what follows we bound each of the above terms individually.  \\

\noindent {\it Bound on (III)}. Let  $ \bar{\theta}^{\prime}{ X}_i $ be a point between  $\widehat{\theta}^{\prime}{ X}_i $ and  ${\theta}^{*\prime}{ X}_i$, for each $i$. A first order Taylor expansion yields
\begin{eqnarray}T_n &=& \frac{1}{2n} \sum_{j=0}^{M}\sum_{i=1}^{2n}X_{ij}X_{ik}\frac{\exp( \bar{\theta}^{\prime}{ X}_i ) }{ (1 + \exp(  \bar{\theta}^{\prime}{ X}_i ) )^2 } (\widehat{\theta}_j \,{-}\, \theta_j^*)\nonumber \\
& = & \frac{1}{2n}\sum_{j=0}^{M}\sum_{i=1}^{2n}X_{ij}X_{ik}\frac{\exp( \bar{\theta}^{\prime}{ X}_i  - \theta^{*\prime}X_i) }{ (1 + \exp(  \bar{\theta}^{\prime}{ X}_i  - \theta^{*\prime}X_i) )^2 } p_1(X_i)p_0(X_i)(\widehat{\theta}_j \,{-}\, \theta_j^*), \nonumber\end{eqnarray}

\noindent where the last line follows from the first by dividing and multiplying the summands by $p_1(X_i)p_0(X_i)$ defined in (\ref{crux}) above. Let  $\gamma$ be arbitrarily close to zero, fixed. Let $D_{\gamma}$ be the set on which 
$\sup_x| \widehat{\theta}^{\prime}x  - \theta^{*\prime}x| \leq \gamma$.  Corollary \ref{firstsup} above guarantees that  $\PP( D_{\gamma}^c )\rightarrow 0$.  Since $\gamma$ is arbitrarily close to zero, the difference  \[ |D_{ni}| =: \left |\frac{\exp( \bar{\theta}^{\prime}{ X}_i  - \theta^{*\prime}X_i) }{ (1 + \exp(  \bar{\theta}^{\prime}{ X}_i  - \theta^{*\prime}X_i) )^2 } -\frac{1}{4}\right | \leq \vartheta, \] with $\vartheta$  arbitrarily close to 0, for each $i$, on the set $D_{\gamma}$.  Therefore we have 
\begin{eqnarray}
 && \PP\left ( |S_n - T_n| \geq \frac{|\beta_k^*| - 2r}{3} \right) \nonumber \\
&&\leq \PP\left (\left| \frac{1}{2n}\sum_{j=0}^{M}\sum_{i=1}^{2n}X_{ij}X_{ik}\left(1 - 
\frac{p_1(X_i)p_0(X_i)}{4}\right)(\widehat{\theta}_j \,{-}\, \theta_j^*)\right|  \geq  \frac{|\beta_k^*| - 2r}{6} \right) \nonumber \\
&&  \ \ +  \ \PP\left ( \left| \frac{1}{2n}\sum_{j=0}^{M}\sum_{i=1}^{2n}D_{ni} X_{ij}X_{ik}(\widehat{\theta}_j \,{-}\, \theta_j^*) \right| \geq  \frac{|\beta_k^*| - 2r}{6} \right) + \PP( D_{\gamma}^c ) \nonumber \\
&&  =: (a) + (b) + \PP( D_{\gamma}^c ).  \nonumber 
\end{eqnarray}

\noindent To bound $ (a) $ we  first invoke {\it Conditions 1 and 2}  to obtain that 
\[ \left| \frac{1}{2n}\sum_{j=0}^{M}\sum_{i=1}^{2n}X_{ij}X_{ik}\left(1 - 
\frac{p_1(X_i)p_0(X_i)}{4}\right)(\widehat{\theta}_j \,{-}\, \theta_j^*)\right|  \leq 
\frac{C}{k^*}\sum_{j=0}^{M}|\widehat{\theta}_j \,{-}\, \theta_j^* |, \]
 where $C$ is a positive constant, independent of $n$ and  depending only on the constant $d$ of {\it Conditions 1 and 2}.  Combining this with the assumption that $\min_{k \in I^*}|\beta_k^*| > 4r$ we arrive at
 \[ (a) \leq \PP\left( \sum_{j=0}^{M} |\widehat{\theta}_j \,{-}\, \theta_j^* | \geq  ck^*r \right) , \]
 \noindent for some positive constant $c$ depending only on $d$ and not on $n$. \\
   
   \noindent  To bound $ (b)$ we  recall that all $X$ variables are  bounded by  $L$ and that $|D_{ni}|$ is bounded by $\vartheta$. Note that  we can always choose $\vartheta = \frac{1}{k^*}$, as we can always choose the corresponding $\gamma$. Therefore, the resulting bound is 
   \[(b) \leq   \PP\left( \sum_{j=0}^{M} |\widehat{\theta}_j \,{-}\, \theta_j^* | \geq  c_1k^*r \right),\]
 \noindent for some $c_1 > 0$ depending on $L$ and not on $n$.  Collecting the bounds above we thus have 
 
 \begin{eqnarray}&& (III) \leq  M\PP\left( \sum_{j=0}^{M} |\widehat{\theta}_j \,{-}\, \theta_j^* | \geq  ck^*r \right) + M \PP\left( \sum_{j=0}^{M} |\widehat{\theta}_j \,{-}\, \theta_j^* | \geq  c_1k^*r \right) \nonumber \\
  &&\hspace{1.1cm}  \ \ \ +  \ M\PP( D_{\gamma}^c ) \nonumber \\
 &&\hspace{1cm}  \leq  3M(\PP(\mathcal{E}^0) + \PP(\mathcal{E}^1)), \nonumber 
 \end{eqnarray}
   where the last line follows from the  proofs of Theorem \ref{rates} and Corollary \ref{firstsup}, for  sets $\mathcal{E}^0, \mathcal{E}^1$  defined as in (\ref{ee}) above,  and with $r$ replaced by the slightly larger value  given in (\ref{newr}) above. Then, as in the proof of {\it Fact 2} of Theorem \ref{delta} and for this value of $r$, we obtain
     \[ (III) \ \leq 3M \frac{2\delta}{M} \leq 6\delta \longrightarrow 0,\]
   \noindent for any $\delta = \delta_n \rightarrow 0$, as $n \rightarrow \infty$. \\
 
\noindent {\it Bound on (II)}. We reason exactly as above,  using now {\it Assumption 2 } and {\it Condition 1} to obtain, for some constant $c > 0$,  that 
\begin{eqnarray}  (II) & \leq & M \PP\left( \sum_{j=0}^{M} |\widehat{\theta}_j \,{-}\, \theta_j^* | \geq  ck^*r \right) \nonumber \\
& \leq & M (\PP(\mathcal{E}^0) + \PP(\mathcal{E}^1)) \leq M \frac{2\delta}{M} \leq 2\delta \longrightarrow 0,
\end{eqnarray}
   \noindent for any $\delta = \delta_n \rightarrow 0$, as $n \rightarrow \infty$. 
\bigskip 

\noindent {\it Bound on (I)}.  First notice that, for any $k$ we have  

\begin{eqnarray}\frac{\partial L_n({\theta^*})}{\partial \beta_k} =
\frac{1}{2n}\left\{ \sum_{i=1}^{n}X_{ik}^{0}p_1(X_i^0) -\sum_{i=1}^{n} { X}_{ik}^1+ 
\sum_{i=1}^{n}X_{ik}^{1}p_1(X_i^1) \right\}. \nonumber
\end{eqnarray}

\noindent Let 

\[ Z_{ik} = X_{ik}^{0}p_1(X_i^0) + X_{ik}^{1}p_1(X_i^1) - { X}_{ik}^1,\]

\noindent and notice that $\EE Z_{ik} = 0$, for all $i$ and $k$.  Since $|Z_{ik}| \leq 3L$, we use Hoeffding's inequality to obtain 

\[\PP\left(\left|\frac{\partial L_n({\theta^*})}{\partial \beta_k} \right| \geq v \right) \leq \exp \left (-\frac{2nv^2}{9L^2}\right),\]
\noindent and this is bounded by $\delta/M$ for any $v \geq \frac{3L\sqrt{\log M/\delta}}{\sqrt{2n}}$. Thus, by {\it Assumption 2} and our choice of $r$ given in (\ref{newr}), we have that 
$ (I) \leq \delta.$ \\

\noindent Collecting now the bounds on $(I), (II)$ and $(III)$ we therefore obtain 
\[ \PP(I^* \not \subseteq \hat{I}) \leq 9\delta \rightarrow 0,\] 
\noindent for any $\delta = \delta_n \rightarrow 0$, as $n \rightarrow \infty$, which completes the first part of the proof.\\

\bigskip 

\noindent {\it Control of $\PP(\widehat{I} \not \subseteq I^*) $.} The control of this quantity will essentially involve the usage of the same probability bounds as above. 
We will however need a different intermediate argument. Let $\mu \in \RR^{k^* + 1}$, where we denote the first component of $\mu$ 
 by $\delta$.  Define 
\begin{eqnarray}H(\mu) &=&  \frac{1}{2n}\left\{ \sum_{i=1}^{2n}\log ( 1 + e^{\mu^{\prime}X_i^0})  -\sum_{i=1}^{n}( \mu^{\prime}X_i^1)+ \sum_{i=1}^{n}\log( 1 + e^{\mu^{\prime}X_i^1} ) \right \} \nonumber \\
&& \ + \ 2r \sum_{j \in I^*}|\mu_j|, \nonumber \end{eqnarray}
and define
\begin{eqnarray}\label{(2.9)}
\widetilde{\mu} = \mathop{\arg \min}_{\mu\in \RR^{k^{*} + 1}} H(\mu),
\end{eqnarray}
\noindent where by convention we denote the first component of $\widetilde{\mu}$ by 
$\widetilde{\delta}$.  Let, by abuse of notation, $\widetilde{\mu} \in \RR^{M +1}$ be the vector that has the first component $\widetilde{\delta}$, the other components of $\widetilde{\mu}$ in positions corresponding to the index set $I^*$,  and is  zero otherwise.
\noindent Define the set 

\[\mathcal{B}_1 = \bigcap_{k \notin I^*}\left\{   \left|\frac{\partial L_n(\widetilde{\mu})}{\partial \beta_k} \right|  < 2r               \right\}.\]

\noindent 
By standard results in convex analysis it follows that, on the set $\mathcal{B}_1$,
$\widetilde{\mu}$ is a solution of (\ref{log}). Recall that $\widehat{\theta}$ is a solution of (\ref{log}) by construction. Then, using simple convex analysis arguments as in  Proposition 4.2 in Bunea (2008), we obtain that  any two solutions have non-zero elements in the same positions. Since, on the set $\mathcal{B}_1$, $\widehat{\theta}_k = 0$ for $k \in {I^*}^c$ we conclude that $\hat{I} \subseteq I^*$ on the set $\mathcal{B}_1$. Hence
\begin{eqnarray}\label{eq:main}
 &&\PP(\hat{I} \not \subseteq I^*) \ \leq \ \PP( \mathcal{B}_1^c) \\
&& \hspace{1.7cm} \leq    M \ \max_{k \notin I^*}  \PP \left ( \left|   \frac{\partial L_n({\theta^*})}{\partial \beta_k}    \right| \geq   r  \right) \nonumber \\
&& \hspace{1.8cm}\ +  \ M \ \max_{k \notin I^*} \PP\left ( \sum_{j = 0}^{M} |\widetilde{\mu}_j - \theta_j^*|
\Bigg|\frac{1}{2n}\sum_{i=1}^{2n}X_{ij}X_{ik}\Bigg| >   r/2   \right) \nonumber \\
&& \hspace{1.8cm} \ + \ M \ \max_{k \notin I^*} \PP\left (\left |S_n(\widetilde{\mu}) - T_n(\widetilde{\mu})\right | \geq r/2 \right), \nonumber 
\end{eqnarray}
\noindent where $T_n$ and $S_n$ have been defined in (\ref{useful}) above.  We notice that the display (\ref{eq:main})  is almost identical to display (\ref{later}) above, and so it can be bounded in a similar fashion. The only difference is in invoking versions of  Theorem \ref{rates} and Corollary \ref{delta} corresponding to $\widehat{\theta}$ replaced by $\widetilde{\mu}$, which hold under the same assumptions and for the same tuning parameters. Consequently
\[ \PP(\hat{I} \not \subseteq I^*) \leq 9\delta \rightarrow 0,\]

\noindent which concludes the proof of this theorem. $\blacksquare$.


\begin{thebibliography}{99}

\bibitem{and72}
{\sc Anderson, J. A.}(1972) Separate sample logistic discrimination. {\em Biometrika}, 
{\bf 59}, 19-35.


\bibitem{bea99}
{\sc Barron, A., Birg\'{e}, L.} and {\sc  Massart, P.} (1999) Risk bounds for model selection via penalization. {\em Probability Theory and Related Fields}, \textbf{113},  301 -- 413.

\bibitem{bresrw00}
{\sc Breslow, N. E., Robins, J. M.} and {\sc Wellner, J. A.} (2000)
On the semi-parametric efficiency of logistic regression under case-control sampling, {\em Bernoulli}, {\bf 6(3)}, 447-455.

\bibitem{bun08}
{\sc Bunea, F.} (2008 ) Honest variable selection in linear and logistic models via 
$\ell_1$ and  $\ell_1 + \ell_2$ penalization, {\em Electronic Journal of Statistics }, {\bf 2,} 1153 - 1194.

\bibitem{buian05}
{\sc Bunea, F.} and {McKeague, I.} (2005) Covariate Selection for Semiparametric Hazard Function Regression Models, {\em Journal of Multivariate Analysis}, {\bf  92}, 186-204. 

\bibitem{bf01}
{\sc Burden, R.L.} and {\sc Faires, J.D.} (2001), Numerical analysis, {\em 7th ed., Pacific Grove, CA: Brooks/Cole}
  
\bibitem{carol95}
{\sc Carroll , R. J., Wang, S.} and {\sc Wang, C. Y.} (1995)
Prospective Analysis of Logistic Case-Control Studies. {\em Journal of the American Statistical Association}, {\bf 90(429)}, 157-169.

\bibitem{chcarnil09}
{\sc Chen,  Y-H., Chatterjee, N.} and {\sc Carroll , R. J.} (2009) Shrinkage Estimators for Robust and Efficient Inference in Haplotype-Based Case-Control Studies. {\em Journal of the American Statistical Association}, {\bf 104(485)},  220Ð233. 

\bibitem{luc}{\sc  Devroye, L.} and { \sc Lugosi, G.} (2001) \textit{Combinatorial methods in density estimation}, Springer-Verlag.

\bibitem{far79}
{\sc Farewell, V. T. }(1979) Some results on the estimation of logistic models based on retrospective data. {\em Biometrika}, {\bf 66}, 27 - 32.

\bibitem{gvw88}
{\sc Gill, R. D., Vardi, Y.}  and {\sc Wellner, J. A.} (1988). Large sample theory of empirical distributions in biased sampling models. {\em Ann. Statist.}  {\bf 16}, 1069-1112. 

\bibitem{h07}
{\sc Hastie, T., Rosset S., Tibshirani, R}, and {\sc Zhu J.} (2004). The Entire Regularization Path for the Support Vector Machine. {\em J. Mach. Learn. Res.} {\bf 5}, 1391-1415.

\bibitem{ksb07}
{\sc Koh, K., Seung-Jean K.}, and {\sc Boyd, S.} (2007) An Interior-Point Method for Large-Scale L1-Regularized Logistic Regression. {\em J. Mach. Learn. Res.} {\bf 8}, 1519-1555.

\bibitem{llw06}
{\sc Leng, C., Lin, Y.}, and {\sc Wahba, G.} (2006). A note on the lasso and related
procedures in model selection. {\em Statistica Sinica} {\bf 16}, 1273-1284.

\bibitem{mgb09}
{\sc Meier, L., van de Geer, S.} and {\sc B\"uhlmann, P.} (2009)
High-dimensional additive modeling, {\em Ann.  Statist., to appear }.

\bibitem{maad01}
{\sc Murphy, S. A.} and {van der Vaart, A. W.} (2001) Semiparametric mixtures in case-control studies, {\em Journal of Multivariate Analysis}, {\bf 79}, 1Ð32. 

\bibitem{os09}
{\sc Osius, G.}  (2009) Asymptotic inference for semiparametric association models,
{\em Ann.  Statist.}, {\bf 37 (1)}, 459-489.

\bibitem{ph06}
{\sc Park, M.Y.} and {\sc Hastie, T.}, (2006). Regularization path algorithms for detecting gene interactions, {\em Manuscript. Available from www-stat.stanford.edu/$\sim$ hastie/Papers/glasso.pdf}.

\bibitem{ph07}
{\sc Park, M.Y.} and {\sc Hastie, T.}, (2007). L1-regularization path algorithm for generalized linear models. {\em Journal of the Royal Statistical Society: Series B}, {\bf 69(4)}, 659-677.

\bibitem{pp79}
{\sc Prentice, R. L.} and {\sc Pyke, R.} (1979)
Logistic disease incidence models and case-control studies. {\em Biometrika}, {\bf 66(3)}, 403 - 411.

\bibitem{qinz}
{\sc Qin, J. } and {\sc Zhang, B.} ( 1997) 
A goodness-of-fit test for logistic regression models based on case-control data
{\em Biometrika}, {\bf 84(3) }, 609-618. 

\bibitem {rwl} {\sc Ravikumar, P.,  Wainwright, M. J.}  and {\sc  Lafferty, J.}  (2008)  High-dimensional graphical model selection using $\ell_1$-regularized logistic regression.  {\em Technical Report, UC Berkeley, Dept of Statistics.} 


\bibitem{r05}
{\sc Rosset, S} (2005). Tracking curved regularized optimization solution paths. {\em Advances in Neural Information Processing Systems} {\bf 17} 2005.

\bibitem{rj07}
{\sc Rosset, S} and {\sc  Zhu, J.} (2007)  Piecewise linear regularized solution paths,  {\em The Ann.  Statist.}, \textbf{35(3)}, 1012-1030.

\bibitem{wa07}
{\sc Shi, W., Lee, K. E.} and {\sc  Wahba, G.} (2007)  Detecting disease-causing genes by LASSO-Patternsearch algorithm. {\em  BMC Proceedings 2007},  {\bf 1(Suppl 1)},  S60.

\bibitem{sara08}{\sc van de Geer, S.} (2008) High-dimensional generalized linear models and the Lasso. {\em The Ann.  Statist.},  \textbf{36(2)}, 614--645.


\bibitem{mart}{\sc Wegkamp, M. H.} (2007) Lasso type classifiers with a reject option. {\em Electronic Journal of Statistics},  {\bf 1}, 155--168.

\bibitem{wu09}
{\sc Wu, T.T., Chen, Y. F., Hastie, T., Sobel, E.} and {\sc  Lange, K.} (2009)
Genome-wide association analysis by lasso penalized logistic regression. {\em Bioinformatics}, {\bf 25(6)}, 714 - 721.


\end{thebibliography}
\end{document}